%% file: main.tex
\documentclass{article}

\usepackage{arxiv}

\usepackage[utf8]{inputenc}
\usepackage[T1]{fontenc}
\usepackage{courier}
\usepackage[hyphens]{url}
\usepackage{graphicx}
\usepackage{natbib}
\usepackage{caption}
\usepackage{booktabs}
\usepackage{amsmath}
\usepackage{amssymb}
\usepackage{mathtools}
\usepackage{amsthm}
\usepackage{amsfonts}
\usepackage{enumitem}
\usepackage{microtype}
\usepackage[colorlinks=true,linkcolor=blue,citecolor=blue,urlcolor=blue]{hyperref}
\input{math_commands.tex}

\title{Weighted Spline-Expanded Networks with Distributional Balancing for Continuous Treatment Effects}

\author{
  Shucheng Liu \\
  University of North Carolina \\
  at Chapel Hill \\
  \texttt{shucheng@unc.edu} \\
  \And
  Chan Park \\
  University of Illinois \\
  Urbana-Champaign \\
  \texttt{parkchan@illinois.edu} \\
  \And
  Guanhua Chen \\
  University of Wisconsin--Madison \\
  \texttt{gchen25@wisc.edu} \\
}

\hypersetup{
  pdftitle={Weighted Spline-Expanded Networks with Distributional Balancing for Continuous Treatment Effects},
  pdfauthor={Shucheng Liu, Chan Park, Guanhua Chen},
}

\begin{document}

\maketitle

\begin{abstract}
  Estimating causal effects with continuous treatments in observational studies is challenging due to confounding, model misspecification, and high-dimensional covariates. We propose the \textit{Weighted Spline-Expanded Network} (WSENet), an end-to-end neural framework that addresses these challenges by combining covariate balancing, structured treatment embedding, and bias-corrected outcome estimation. WSENet first applies \textit{Distance Covariate Optimal Weights} to induce distributional independence between covariates and treatment without relying on parametric models. It then learns the conditional outcome via a structured network that fuses outcome-relevant representations of covariates with a spline-expanded treatment input, enabling smooth and flexible modeling of the dose-response relationship. To mitigate residual bias, we introduce \textit{Weighted Targeted Regularization}, a correction technique based on efficient influence functions that yields a doubly robust estimator. Extensive evaluations on semi-synthetic and real-world datasets, including high-dimensional genomic and environmental health data, demonstrate that WSENet consistently outperforms existing baselines in both accuracy and stability.
\end{abstract}

\section{Introduction}

Estimating causal effects for continuous treatments is crucial in 
domains such as medicine (e.g., drug dosages) \citep{schweisthal2023reliable, 
chakraborty2014dynamic}, economics (e.g., income levels) 
\citep{pickett2015income}, and environmental science (e.g., air pollution 
indices) \citep{dominici2022assessing, imai2004causal}. A key quantity in 
these studies is the \textit{Average Dose-Response Function (ADRF)} 
\citep{imai2004causal, kennedy2017non, bahadori2022end, wang2022generalization, 
gao2024variational,campana2024predicting,kazemi2024adversarially}, which 
represents the expected outcome at each treatment level. Accurate ADRF 
estimation enables practitioners to evaluate treatment efficacy, optimize 
intervention strategies \citep{li2025reinforcement, cai2023jump}, and inform 
policy learning for more effective decision-making \citep{kallus2018policy, 
chernozhukov2019semi, schweisthal2023reliable, galvao2015uniformly, 
qi2023robustness}.

Due to the high costs of randomized controlled trials (RCTs), researchers 
often rely on observational studies to estimate the ADRF \citep{schwab2020learning, 
Hu2024DTRNetPC, bahadori2022end, li2023quasi}, where confounding poses the 
primary challenge \citep{bareinboim2012controlling, hernan2004structural, 
galvao2015uniformly}. The generalized propensity score (GPS) 
\citep{inbook} adjusts for confounding by modeling the 
conditional treatment density, but is sensitive to model misspecification 
and often requires unstable inverse weighting \citep{kallus2018policy}. 
Weighting methods such as CBPS \citep{imai2014covariate} and 
IPM-based approaches \citep{kong2023covariate} estimate balancing weights 
more directly, yet may suffer from identifiability issues or poor scaling 
in high-dimensional settings. Balancing alone, however, is insufficient 
for efficient ADRF estimation and must be coupled with flexible outcome 
modeling. Parametric outcome models impose restrictive functional assumptions 
\citep{guardabascio2014estimating, imbens2000role}, while nonparametric 
methods such as kernel regression \citep{kallus2018policy, cai2021deep} 
can exhibit high variance in data-sparse regions. These limitations motivate 
the development of methods that are both flexible and stable for continuous 
treatment settings.

Recent neural network approaches have improved flexibility under 
high-dimensional covariates, but introduce their own compromises. 
DRNet \citep{schwab2020learning} discretizes the treatment into bins, 
sacrificing ADRF smoothness, while VCNet \citep{nie2021vcnet} avoids 
discretization but relies on grid-based treatment sampling and explicit 
GPS inversion, adding computational overhead and instability in 
high-dimensional regimes. More broadly, since we only require balancing 
weights rather than the full treatment density for bias correction, 
explicitly modeling an entire GPS function adds unnecessary complexity. 
A more direct approach to confounding adjustment is desirable.

To estimate the ADRF under continuous treatments with high-dimensional 
covariates, we propose the Weighted Spline-Expanded Network (WSENet), an 
end-to-end neural framework that addresses these challenges through three 
matched components. We use neural networks to learn low-dimensional 
covariate representations predictive of outcomes, and expand the treatment 
variable using spline basis functions \citep{prichard1971assessment, 
threlfall1999sun} to flexibly capture nonlinear dose-response relationships 
without discretization. To adjust for confounding without relying on 
propensity score modeling, we incorporate DCOW \citep{huling2024independence}, 
a distributional balancing method that directly minimizes dependence between 
covariates and treatment, integrated into a weighted loss function for 
stable, model-free confounding adjustment. To further reduce residual bias 
of the plug-in estimator, we introduce Weighted Targeted Regularization 
(WTR), which adjusts the network's predictions toward satisfying causal 
estimating equations derived from the efficient influence function (EIF), 
resulting in a doubly robust estimator that remains consistent when either 
the outcome model or the weighting is correctly specified \citep{van2011targeted}. 
Experiments on semi-synthetic and real-world datasets confirm that WSENet 
consistently delivers superior ADRF estimation accuracy, particularly in 
high-dimensional covariate settings.

Our contributions are threefold. We introduce a DCOW-driven weighted 
outcome learning framework that achieves stable, model-free deconfounding 
without estimating or inverting the GPS. We propose a spline-expanded 
neural architecture integrated with EIF-based weighted targeted 
regularization, enabling flexible ADRF estimation with doubly robust 
guarantees. Extensive experiments on semi-synthetic and real-world datasets 
demonstrate that WSENet consistently outperforms state-of-the-art 
kernel-based and deep learning baselines in both accuracy and stability.

\section{Related Work}
\textbf{Weighting and Kernel-Based ADRF Estimation.} Early work on continuous treatments extended the propensity score framework to the dose-response setting. \citet{inbook} introduced the GPS and showed that conditioning on it identifies the ADRF, but parametric GPS estimation is sensitive to density misspecification and inverse weighting can be highly unstable \citep{kallus2018policy}. To circumvent direct density estimation, \citet{imai2014covariate} proposed CBPS, which selects weights by moment-balancing rather than likelihood, while \citet{kong2023covariate} cast covariate balancing as minimizing an integral probability metric. \citet{huling2024independence} introduced DCOW, which directly minimize a weighted distance covariance to enforce distributional independence between covariates and treatment, providing model-free and assumption-light deconfounding. On the outcome side, kernel-smoothing estimators \citep{kallus2018policy, cai2021deep} provide nonparametric flexibility but tend to inflate variance in sparse or high-dimensional regions.

\textbf{Deep Learning for Continuous Treatment Effects.} Neural approaches aim to handle high-dimensional covariates and nonlinear dose-response surfaces. DRNet \citep{schwab2020learning} discretizes the treatment interval and fits separate outcome heads per bin, sacrificing ADRF smoothness. VCNet \citep{nie2021vcnet} avoids discretization through spline-based varying coefficients but jointly models a GPS branch and evaluates targeted regularization on a treatment grid, introducing significant computational overhead and stability issues under high-dimensional covariates. Several recent variants refine these designs: SCIGAN \citep{bica2020estimating} uses a hierarchical GAN to generate counterfactual outcomes across dosages; ACFR \citep{kazemi2024adversarially} employs adversarial cross-attention representations; ADMIT \citep{wang2022generalization} learns a reweighting network to alleviate selection bias; and KernelNN \citep{colangelo2025double} couples a kernel neural estimator with a multi-GPS density model for doubly debiased inference. Despite their diversity, these methods either commit to estimating and inverting the GPS or rely on treatment discretization.

\textbf{Doubly Robust Estimation and Targeted Regularization.} Doubly robust methodology originates with augmented inverse probability weighting \citep{robins1994estimation, bang2005doubly}, which retains consistency if either the outcome model or the weighting model is correctly specified. \citet{kennedy2017non} extended this to continuous treatments and derived nonparametric doubly robust estimators of the ADRF. Efficient influence functions \citep{hines2022demystifying, fisher2021visually} provide the theoretical foundation for constructing such estimators, and Targeted Maximum Likelihood Estimation (TMLE) \citep{van2011targeted} operationalizes EIF-based corrections within a likelihood framework. Within neural pipelines, \citet{shi2019adapting} first incorporated targeted regularization as an end-to-end training penalty for binary treatments; VCNet-TR \citep{nie2021vcnet} adapted this idea to continuous treatments but ties the correction to an explicit GPS estimated on a treatment grid.

\section{Problem Setting}
We assume an independent and identically distributed (i.i.d.) dataset \( \{(Y_i,\boldsymbol{X}_i,T_i)\}_{i=1}^n \) which are sampled as vectors \( (\boldsymbol{X}_i, T_i, Y_i) \), where \( \boldsymbol{X}_i \in \mathbb{R}^p \) is a vector of covariates, potentially high-dimensional, \( T_i \in \mathbb{R} \) represents continuous treatment, and \( Y_i \in \mathbb{R} \) is the outcome. For convenience, we normalize \( T \) to the range \([0, 1]\). Within the potential outcome framework, our goal is to estimate the Average Dose Response Function (ADRF) under continuous treatment 
$
\varphi (t) = \mathbb{E}[Y(t)].
$
The ADRF can be identified from observational data under the following assumptions \citep{miguel2023causal, hernan2010causal, rosenbaum1983central}.

\noindent\textbf{Assumption 1 (Stable Unit Treatment Value Assumption (SUTVA))}: There are no interactions between units, and each treatment level has only one version. Different doses or levels of a treatment are considered distinct treatments, then $Y = Y(T)$.

\noindent\textbf{Assumption 2 (Ignorability)}: The potential outcome \( Y(t) \) is independent of the assignment of treatment given all covariates, that is, there is no unobserved confounding. Formally, $Y(t) \perp T \mid \boldsymbol{X}$.

\noindent\textbf{Assumption 3 (Positivity)}: Each unit must have a non-zero probability of being assigned to each treatment level. Formally, $f(T = t \mid \boldsymbol{X} = x) > c$ for some $c>0$, $\forall t \in [0, 1], \, x \in \mathcal{X}$.

Under these assumptions, we have
\begin{align}
\varphi(t) &= \mathbb{E}[\mathbb{E}(Y(t)|\mathbf{X})] = \mathbb{E}[\mathbb{E}(Y(t)|\mathbf{X}, T=t)] \nonumber \\
&= \mathbb{E}[\mathbb{E}(Y|\mathbf{X}, T=t)] = \mathbb{E}[\mu( \mathbf{X}, t)]. \nonumber
\end{align}

These assumptions are standard in the ADRF literature and are shared by essentially all competing methods; when they are questionable, we discuss diagnostics and remedies in the Conclusion and in the supplementary material. Existing estimators fall into two families. Outcome-modeling methods regress $Y$ on $\mathbf{X}$ and $t$ to estimate $\mu(\mathbf{X},t)=\mathbb{E}[Y\mid\mathbf{X},T=t]$ and average over the covariate distribution, $\varphi(t)=\mathbb{E}_{\mathbf{X}}[\mu(\mathbf{X},t)]$; they are efficient when $\mu$ is correctly specified but biased under misspecification, especially with confounding or high-dimensional $\mathbf{X}$. Weighting methods instead reweight samples by the generalized propensity score (GPS) $f(T\mid\mathbf{X})$ \citep{inbook} so that covariates balance across treatment values, which avoids modeling $\mu$ but requires estimating $f(T\mid\mathbf{X})$ and produces unstable, high-variance estimates when some units have near-zero conditional density.

Given these complementary strengths and weaknesses, we combine both in a flexible weighted outcome modeling approach. We first learn balancing weights that mitigate the statistical dependence between $\mathbf{X}$ and $T$ on the weighted scale, then integrate these weights into the outcome network. To further reduce bias, we apply a weighted targeted regularization strategy during training. This design yields a doubly robust and efficient ADRF estimator, as detailed below.

\section{Methodology}
This section introduces our framework for estimating ADRF under continuous treatments. Our method, termed the \textit{Weighted Spline-Expanded Network (WSENet)}, combines the strengths of outcome modeling and weighting-based adjustment to achieve both robustness and flexibility in high-dimensional settings. 

We begin by estimating balancing weights in a robust, assumption-light manner using \textit{Distance Covariate Optimal Weights (DCOW)}, which optimize a dependence-minimizing distance metric to render the covariates \( \mathbf{X} \) independent of the treatment \( T \). These weights mitigate confounding without relying on parametric models for the treatment mechanism, avoiding instability from extreme inverse propensity weights. We then integrate the learned weights into a neural outcome model designed to estimate \( \mu(\mathbf{X}, t) \), the conditional mean outcome given covariates and treatment. To capture nonlinear dose-response relationships, we expand the continuous treatment variable \( t \) using spline basis functions and feed both the spline-expanded treatment and covariate representations into the network. This design enables the model to flexibly learn a smooth ADRF, while incorporating sample-specific weights to correct for covariate imbalance. To further reduce residual bias from the plug-in estimator, we introduce a \textit{weighted targeted regularization} (WTR) procedure, which augments the training loss with a bias correction term. This final step yields a doubly robust estimator: consistent if either the outcome model or the weighting model is correctly specified. In the following subsections, we provide detailed explanations of WSENet with the structure illustrated in Figure~\ref{fig:sample}.

\subsection{Distance Covariate Optimal Weights for inducing independence and deconfounding}
Before describing our network structure with a weighted loss, we first introduce how to estimate the weights that induce independence and correct for confounding. Our primary objective is to utilize these weights to force the weighted joint empirical distribution to approximate the product of their unweighted marginals, expressed as 
$F_{\mathbf{X}, T, \mathbf{w}}^n \approx F_{\mathbf{X}}^n F_T^n,$
where \( \mathbf{w} = (w_1, \ldots, w_n) \) is a vector of weights such that \( \sum_{i=1}^n w_i = n \) and \( w_i \geq 0 \) for all \( i = 1, \ldots, n \). Here, \( F_{\mathbf{X}}^n = n^{-1} \sum_{i=1}^n I(\mathbf{X}_i \leq \mathbf{x}) \) denotes the empirical cumulative distribution function (CDF) of \( \{\mathbf{X}_i\}_{i=1}^n \), \( F_T^n = n^{-1} \sum_{i=1}^n I(T_i \leq t) \) is the empirical CDF of \( \{T_i\}_{i=1}^n \), and \( F_{\mathbf{X}, T, \mathbf{w}}^n(\mathbf{x}, t) = n^{-1} \sum_{i=1}^n w_i I(\mathbf{X}_i \leq \mathbf{x}, T_i \leq t) \) represents the weighted empirical CDF of \( \{\mathbf{X}_i, T_i\}_{i=1}^n \) using weights \( \mathbf{w} \). We use DCOW, a robust, assumption-free method that minimizes the dependence between covariates \( \mathbf{X} \) and treatment \( T \) via a weighted distance covariance objective \citep{huling2024independence}. We define
\[
\mathcal{D}(\mathbf{w}) = \mathcal{V}_{n,\mathbf{w}}^2(\mathbf{X}, T) + \mathcal{E}(F_{\mathbf{X},\mathbf{w}}^n, F_{\mathbf{X}}^n) + \mathcal{E}(F_{T,\mathbf{w}}^n, F_T^n),
\]
where \( \mathcal{V}_{n,\mathbf{w}}^2(\mathbf{X}, T) \) quantifies joint dependence via weighted distance covariance, and the two \( \mathcal{E} \) terms are energy distances that measure how well the marginal covariate and treatment distributions are preserved after weighting; the exact expressions of each term in terms of empirical characteristic functions are provided in the supplementary material. Minimizing \( \mathcal{D}(\mathbf{w}) \) explicitly minimizes the dependence between \( \mathbf{X} \) and \( T \) on the weighted scale while keeping the weighted marginal distributions faithful to the original data. The optimal weights are obtained by solving
\[
\mathbf{w}_n^p \in \underset{\mathbf{w} \in \mathcal{W}_n}{\arg\min} \; \mathcal{D}(\mathbf{w}), 
\]
where $ \mathcal{W}_n = \left\{ \mathbf{w} \in \mathbb{R}^n : \sum_{i=1}^n w_i = n, \; w_i \geq 0 \right\},$ which can be formulated as a quadratic programming problem. Compared to GPS methods, which rely on potentially misspecified parametric models and require inverse density estimation, DCOW directly optimize covariate balance without imposing distributional assumptions. This results in more stable and interpretable weights, especially in high-dimensional settings.

\subsection{Network Structure of the Weighted Spline-Expanded Network}

Once we obtain weights that mitigate confounding by minimizing dependence between treatment \( T \) and covariates \( \boldsymbol{X} \), we focus on modeling the conditional outcome function \( \mu(\boldsymbol{x}, t) = \mathbb{E}(Y \mid  \boldsymbol{X} = \boldsymbol{x}, T = t) \). Neural networks offer a flexible framework for this task, especially under high-dimensional covariates. However, naively embedding \( T \) and \( \boldsymbol{X} \) as input to a feedforward network can obscure the influence of the scalar treatment variable \( t \), especially when \( \boldsymbol{X} \) lies in a high-dimensional space.

To address this, we design WSENet to model treatment and covariates through distinct yet interacting components. The network has two main design elements: (1) outcome-relevant representation learning from covariates and (2) spline-based expansion of the treatment variable. These are combined through a cohesive, end-to-end neural framework that enables flexible and stable estimation of the ADRF.

We employ a neural encoder (a two-layer MLP) to learn a representation \( \tilde{\boldsymbol{X}} \) from high-dimensional \( \boldsymbol{X} \). This latent encoding is a product of the learning process rather than an architectural constraint; it emerges as the network is optimized to extract features most relevant for predicting the outcome \( Y \), reducing variance and enhancing generalization.

We model the treatment variable \( t \) using a B-spline basis expansion \( \phi(t) = (N_{1,d}(t), \ldots, N_{m,d}(t))^\top \in \mathbb{R}^m \), where \( N_{j,d}(\cdot) \) are spline basis functions of degree \( d \) and \( m \) is the number of basis functions determined by the number of knots and spline order. This treatment expansion ensures smoothness and flexibility in modeling nonlinear dose-response effects.

The treatment basis is linearly embedded using a learnable parameter matrix \( B_1 \in \mathbb{R}^{l \times m} \), resulting in the transformed treatment representation \( \eta(t) = B_1 \phi(t) \). A separate bias term \( B_2 \in \mathbb{R}^{1 \times m} \) is also included. The covariate and treatment representations interact through a scalar index function, with the interaction term defined as
\[\text{ReLU}\left( \tilde{\boldsymbol{X}}^\top \eta(t) + B_2 \phi(t) \right),\]
which produces a one-dimensional summary that captures both covariate-treatment interactions and treatment-specific heterogeneity. This structure ensures that the network captures smooth variation in \( t \) while retaining flexibility in \( \boldsymbol{X} \).

\begin{figure*}[t]
    \centering
    \includegraphics[width=15.22cm, height=6cm]{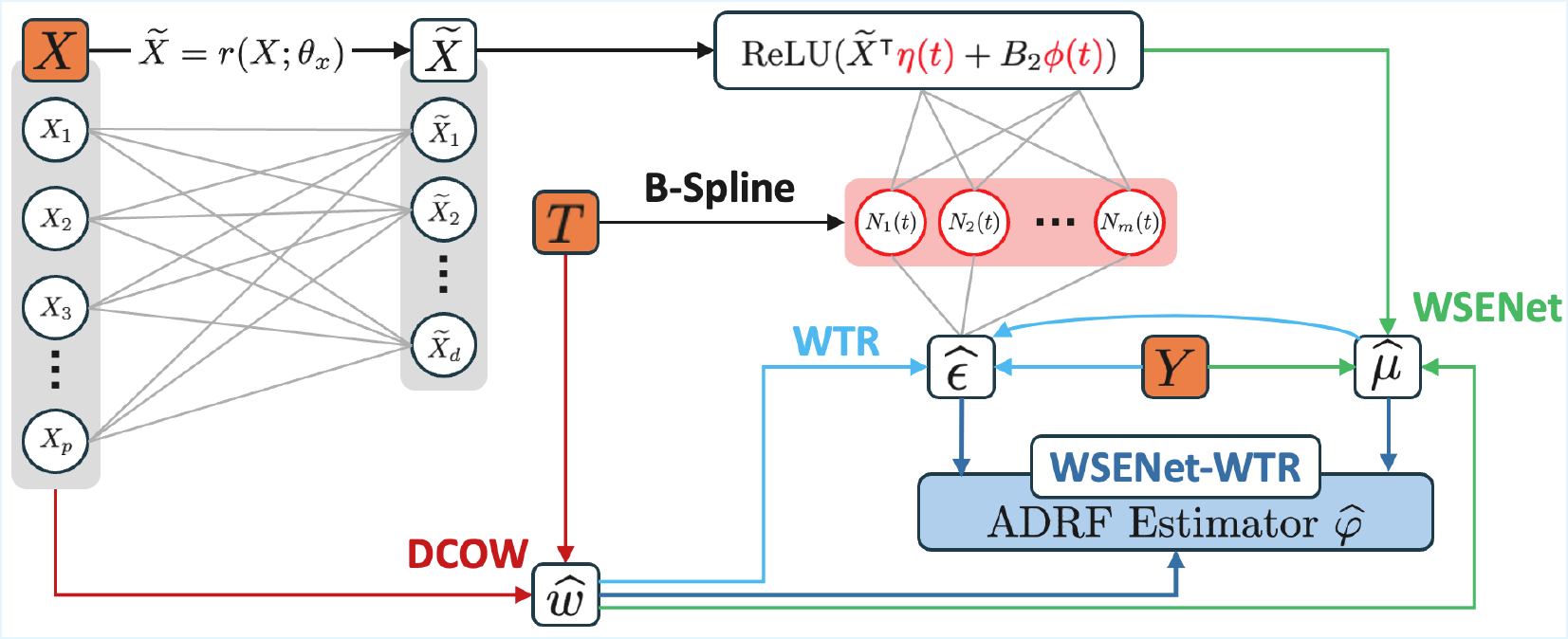}
    \caption{Architecture of WSENet and WSENet-WTR. WSENet embeds DCOW weights into the neural loss and fuses spline-expanded treatments with covariate embeddings, while the WTR variant further incorporates weighted targeted regularization to enhance robustness of ADRF estimation.}
    \label{fig:sample}
\end{figure*}

Additional layers can be stacked to capture higher-order interactions, and the final output is the network's prediction \( \hat{\mu}(t, \boldsymbol{x}) \). This architecture is jointly trained end-to-end with the weighted loss and, optionally, with targeted regularization. Compared to designs like DRNet or VCNet, WSENet avoids discretization of the treatment and integrates balancing weights directly, resulting in smoother and more robust ADRF estimation. A schematic of the architecture is shown in Figure~\ref{fig:sample}.

\subsection{Loss Function Design}
While WSENet captures complex treatment-covariate interactions through its structured architecture, training the model using standard unweighted empirical risk yields biased estimates of the ADRF in observational settings. Strictly speaking, since adjusting for $\mathbf{X}$ theoretically accounts for confounding, this estimation error is primarily driven by the covariate shift of $T|\mathbf{X}$. Specifically, because of unbalanced treatment assignments and poor overlap, an unweighted loss biases the learned conditional mean $\hat{\mu}(\boldsymbol{x}, t)$ toward high-density regions of the observed data, exacerbating the finite-sample bias of the plug-in estimator in sparse $(\boldsymbol{x}, t)$ regions. To correct for this covariate shift and ensure the model accurately reflects the causal effect of $t$, we incorporate balancing weights derived from DCOW.

Using these weights \( \hat{w}_i \), we define the weighted loss function for training WSENet
$$L = \sum_{i=1}^n \left( y_i - \hat{\mu}(\boldsymbol{x}_i, t_i) \right)^2 \hat{w}_i,$$
where \( \hat{\mu}(\boldsymbol{x}_i, t_i) \) is the network's prediction for individual \( i \). This objective function emphasizes samples that are most informative under the reweighted, pseudo-randomized distribution and downweights those in regions of covariate-treatment imbalance.

After training, the marginal ADRF is estimated with the plug-in approach
$
\widehat{\varphi}(t) = \frac{1}{n} \sum_{i=1}^n \hat{\mu}(\boldsymbol{x}_i, t).
$
While this estimator is consistent under correct model specification and sufficient sample size, it may exhibit residual bias in finite samples or when either the outcome model or the weights are imperfect. To further enhance robustness, we propose a bias-corrected estimator using weighted targeted regularization, described next.

\subsection{Weighted Targeted Regularization}

While the plug-in estimator \( \widehat{\varphi}(t) = \frac{1}{n} \sum_{i=1}^n \hat{\mu}(\boldsymbol{x}_i, t) \) is consistent under ideal conditions, it may suffer from bias in small samples or when the model for \( \mu(\boldsymbol{x}, t) \) is misspecified. To address this, we incorporate ideas from semiparametric theory, specifically, the efficient influence function (EIF) \citep{hines2022demystifying, fisher2021visually}, to guide a bias-corrected estimation procedure.

The EIF characterizes the most efficient (i.e., lowest variance) regular estimator of a functional under a nonparametric model. In our case, the integrated ADRF functional \( \psi = \int\!\!\int \mu(\boldsymbol{x}, t) f(\boldsymbol{x}) f(t)\, dt\,d\boldsymbol{x} \) can be used to derive doubly robust estimators \citep{van2011targeted, kennedy2017non, fisher2021visually}; the proof is provided in the supplementary material. Letting \( \mu \) denote the outcome regression and \( w(\boldsymbol{X}, T) = f(\boldsymbol{X}) f(T)/f(\boldsymbol{X}, T) \) denote the true inverse joint density ratio, according to \citet{kennedy2017non} as well as similar derivation in the supplementary material, the main term of EIF for $\psi$ denoted as $\xi(\boldsymbol{X}, Y, T; w, \mu)$ is defined as 
\[   \left(Y - \mu(\boldsymbol{X}, T) \right)w(\boldsymbol{X}, T) + \int \mu(\boldsymbol{x}, T )f(\boldsymbol{x})d\boldsymbol{x}. \]
When we plug in $\hat{\mu}(\boldsymbol{X}, T)$ and $\hat{w}(\boldsymbol{X}, T)$ for $\mu(\boldsymbol{X}, T)$ and $w(\boldsymbol{X}, T)$, then the first term of EIF
represents the residual bias under imperfect outcome modeling. This motivates a correction term for plug-in estimators.

To empirically implement this EIF-based correction within a neural framework, we incorporate this correction into our ADRF estimator using a learned augmentation term. We define a perturbed outcome model
\[
\tilde{\mu}(\boldsymbol{x}, t) = \hat{\mu}(\boldsymbol{x}, t) + \hat{\epsilon}(t)\hat{w},
\]
where \( \hat{\epsilon}(t) \) is a flexible function modeled with a B-spline basis
\[
\hat{\epsilon}(t) = A \phi(t), \quad \phi(t) = (N_{1,d}(t), \ldots, N_{m,d}(t))^\top,
\]
with an extra learnable parameter matrix \( A \in \mathbb{R}^{1 \times m} \), spline degree \( d \), and \( m \) basis functions. This augmentation term is trained jointly within the WSENet architecture. Then, the revised loss function becomes
\[
L = \sum_{i=1}^n \left( y_i - \hat{\mu}(\boldsymbol{x}_i, t_i) - \hat{\epsilon}(t_i) \hat{w}_i \right)^2 \hat{w}_i,
\]
which targets the residual bias via the EIF-based correction. The final ADRF estimator is
\[
\widehat{\varphi}(t) = \frac{1}{n} \sum_{i=1}^n \left( \hat{\mu}(\boldsymbol{x}_i, t) + \hat{\epsilon}(t) \hat{w}_i \right).
\]
This estimator is \textit{doubly robust}, it remains consistent if either the outcome model \( \mu \) or the balancing weights \( w \) are correctly specified. Compared to VCNet-TR \citep{nie2021vcnet}, our approach avoids reliance on inverse propensity scores with grid based treatment, and directly leverages EIF structure for principled and stable bias correction.

\subsection{Theoretical Property of the estimator of WSENet}
To characterize the asymptotic behavior and convergence rate of the proposed estimator \( \varphi(t) \), we first introduce a set of standard regularity assumptions.

\noindent\textbf{Assumption 1:} \( Y = \mu(\mathbf{X}, t) + \eta \), where \( \mathbb{E}[\eta] = 0 \), \( \eta \perp (\mathbf{X},T) \), and \( \eta \) follow a sub-Gaussian distribution.

\noindent\textbf{Assumption 2:} The functions \( w \), \( \mu \), \( \hat{\mu} \), and \( \hat{w} \) have  second derivatives in the functional spaces \( w, \hat{w} \in \mathbb{Q} \), \( \mu, \hat{\mu} \in \mathbb{U}\). Moreover, \( \hat{w} \rightarrow w \) or \( \hat{\mu} \rightarrow \mu \) as $n \rightarrow \infty$ in $L_\infty$ norm. \( \text{Rad}_n(\mathbb{Q}) = O_p(n^{-1/2}),      \ \text{Rad}_n(\mathbb{U}) = O_p(n^{-1/2}) \), where \text{Rad} represent the Rademacher complexity.

\noindent\textbf{Assumption 3:} \( B_{K_n} \) involved in $\phi(t)$ is a closed linear basis of B-splines with equally spaced knots, and \( K_n \asymp n^{1/6} \).

\noindent\textbf{Theorem 1:}
Given the estimator \( \hat{\varphi}(t) = \frac{1}{n} \sum_{i=1}^{n} \left( \hat{\mu}(\boldsymbol{x}_i, t) + \hat{\epsilon}(t) \hat{w}_i \right) \), we have the following asymptotic result
\[
\|\hat{\varphi}(t) - \varphi(t)\|_{L^2} = O_p\left(n^{-\frac{1}{3}} \sqrt{\log n} + r_1(n) r_2(n)\right),
\]
where \( r_1(n) \) and \( r_2(n) \) are the convergence rates of \( \hat{\mu} \) and \( \hat{w} \) in the $L_{\infty}$  norm, respectively. The theorem establishes consistency under mild conditions and shows that the estimation error decomposes into approximation errors from the outcome and weighting components (see the proof in the supplementary material).

\section{Experiments}
To evaluate the performance of our method, we consider three semi-synthetic datasets with varying dimensionality: IHDP \citep{hill2011bayesian}, News \citep{newman2008bag}, and The Cancer Genome Atlas (TCGA) \citep{weinstein2013cancer}. 

\textbf{Datasets.} The IHDP dataset is derived from a randomized experiment on early childhood interventions and includes 747 observations with 25 covariates describing participants (e.g., birth weight, head circumference, preterm birth). We select subsets of 10 and 25 covariates for analysis. The News dataset consists of 3000 New York Times articles represented by 500 covariates based on word frequencies; we evaluate model performance on subsets with 100, 300, and 500 covariates. The TCGA dataset contains gene expression profiles of cancer patients, with 9659 samples and 4000 genes; we use subsets with 1000 and 4000 covariates. Since the true ADRF is unobservable in real-world data, we construct semi-synthetic outcomes by explicitly defining the functional relationships between covariates, treatment, and outcomes, and simulate responses using Monte Carlo methods \citep{rubinstein2016simulation}. Detailed data generation procedures are provided in the supplementary material.

\begin{table*}[t]
  \centering
  \renewcommand{\arraystretch}{1.5}
  \caption{Integrated root mean squared error of different methods under three semi-synthetic datasets.}
  \resizebox{\textwidth}{!}{%
  \begin{tabular}{lcccccccccccccc}
    \toprule
    \textbf{Dataset} & \multicolumn{4}{c}{\textbf{IHDP}} & \multicolumn{6}{c}{\textbf{News}} & \multicolumn{4}{c}{\textbf{TCGA}} \\
    \cmidrule(lr){1-1} \cmidrule(lr){2-5} \cmidrule(lr){6-11} \cmidrule(lr){12-15}
    \textbf{Num} & \multicolumn{2}{c}{\textbf{200}} & \multicolumn{2}{c}{\textbf{500}} & \multicolumn{3}{c}{\textbf{1000}} & \multicolumn{3}{c}{\textbf{2000}} & \multicolumn{2}{c}{\textbf{3000}} & \multicolumn{2}{c}{\textbf{6000}} \\
    \cmidrule(lr){1-1} \cmidrule(lr){2-5} \cmidrule(lr){6-11} \cmidrule(lr){12-15}
    \textbf{Covariates} & \textbf{10} & \textbf{25} & \textbf{10} & \textbf{25} & \textbf{100} & \textbf{300} & \textbf{500} & \textbf{100} & \textbf{300} & \textbf{500} & \textbf{1000} & \textbf{4000} & \textbf{1000} & \textbf{4000} \\
    \cmidrule(lr){1-1} \cmidrule(lr){2-5} \cmidrule(lr){6-11} \cmidrule(lr){12-15}
    GPS         & 1.39  & 0.80  & 1.30  & 0.78  & 0.102 & 0.106 & 0.190 & 0.089 & 0.093 & 0.092 & 2.07e-02 & 1.62e-02 & 1.81e-02 & 1.35e-02 \\
    CBPS        & 1.29  & 0.73  & 1.23  & 0.62  & 0.101 & 0.105 & 0.106 & 0.089 & 0.093 & 0.087 & 2.57e-02 & 2.47e-02 & 2.52e-02 & 2.33e-02 \\
    GBM         & 1.32  & 0.73  & 1.27  & 0.72  & 0.101 & 0.101 & 0.100 & 0.085 & 0.087 & 0.087 & 2.45e-02 & 2.46e-02 & 2.44e-02 & 2.24e-02 \\
    DCOW        & 1.30  & 0.73  & 1.23  & 0.63  & 0.095 & 0.105 & 0.084 & 0.111 & 0.110 & 0.083 & 2.54e-02 & 1.54e-02 & 1.65e-02 & 1.21e-02 \\
    \cmidrule(lr){1-1} \cmidrule(lr){2-5} \cmidrule(lr){6-11} \cmidrule(lr){12-15}
    SCIGAN      & 1.71  & 1.52  & 1.67  & 1.33  & 0.266 & 0.226 & 0.205 & 0.262 & 0.163 & 0.145 & 2.49e-02 & 2.44e-02 & 2.47e-02 & 2.22e-02 \\
    DRNet       & 0.88  & 1.16  & 0.78  & 0.79  & 0.157 & 0.152 & 0.150 & 0.125 & 0.136 & 0.130 & 5.42e-02 & 5.39e-02 & 5.37e-02 & 5.38e-02 \\
    ACFR        & 0.81  & 0.76  & 0.80  & 0.75  & 0.123 & 0.100 & 0.099 & 0.111 & 0.099 & 0.098 & 5.20e-03 & 4.80e-03 & 5.11e-03 & 4.72e-03 \\
    ADMIT       & 1.00  & 0.90  & 0.50  & 0.43  & 0.157 & 0.105 & 0.084 & 0.084 & 0.079 & 0.075 & 8.08e-03 & 7.60e-03 & 5.25e-03 & 3.85e-03 \\
    KernelNN-DR & 0.76  & 0.55  & 0.50  & 0.42  & 0.102 & 0.101 & 0.090 & 0.092 & 0.088 & 0.080 & 1.78e-02 & 1.03e-02 & 1.22e-02 & 8.32e-03 \\
    VCNet       & 0.56  & 1.18  & 0.31  & 0.75  & 0.160 & 0.120 & 0.167 & 0.114 & 0.117 & 0.098 & 1.28e-02 & 1.15e-02 & 1.27e-02 & 9.20e-03 \\
    VCNet-TR    & 0.49  & 0.94  & 0.27  & 0.42  & 0.129 & 0.193 & 0.199 & 0.072 & 0.166 & 0.104 & 4.80e-03 & 9.50e-03 & 5.40e-03 & 8.80e-03 \\
    \cmidrule(lr){1-1} \cmidrule(lr){2-5} \cmidrule(lr){6-11} \cmidrule(lr){12-15}
    WSENet      & 0.49  & 0.96  & 0.31  & 0.88  & 0.160 & 0.150 & 0.090 & 0.112 & 0.135 & 0.144 & \textbf{1.70e-03} & 3.70e-03          & \textbf{2.00e-03} & 3.40e-03          \\
    WSENet-WTR  & \textbf{0.34} & \textbf{0.29} & \textbf{0.17} & \textbf{0.18} & \textbf{0.083} & \textbf{0.074} & \textbf{0.078} & \textbf{0.065} & \textbf{0.049} & \textbf{0.058} & 4.00e-03 & \textbf{2.00e-03} & 4.10e-03          & \textbf{2.10e-03} \\
    \bottomrule
  \end{tabular}%
  }
  \label{tab:table1_final_mean}
\end{table*}

\textbf{Baselines.} Our proposed methods include WSENet and WSENet-WTR, which correspond to versions without and with weighted targeted regularization, respectively. We compare our WSENet framework with state-of-the-art kernel-based and deep learning methods for continuous treatment causal inference. Among kernel-based methods, we use the normalized weighted Nadaraya-Watson (NW) estimator \citep{huling2024independence}
$$
\widehat{\varphi}(t) = \frac{\sum_{i=1}^n Y_i w_i K_h(T_i - t)}{\sum_{i=1}^n w_i K_h(T_i - t)},
$$
where \( w_i \) denotes balancing weights and \( h \) is a bandwidth selected by cross-validation. Weighting methods include: (1) Generalized Propensity Score (GPS) \citep{inbook}, using kernel density estimation and linear modeling to estimate stabilized scores; (2) Covariate Balancing Propensity Score (CBPS) \citep{imai2014covariate}, which optimizes moment balance; (3) Gradient Boosting Machine (GBM) \citep{friedman2001greedy}, which estimates conditional densities using boosted trees; and (4) DCOW \citep{huling2024independence}, which directly learn balancing weights without modeling treatment density. 

Among deep learning baselines, we include SCIGAN \citep{bica2020estimating}, which is a modified generative adversarial network (GAN) framework to estimate counterfactual outcomes;  DRNet \citep{schwab2020learning}, which fits separate heads for discretized treatment levels; ACFR \citep{kazemi2024adversarially}, which utilizes an adversarial and cross-attention network to predict potential outcomes; ADMIT \citep{wang2022generalization}, which learned a re-weighting network aiming to alleviate the selection bias; KernelNN-DR \citep{colangelo2025double} implements a continuous-treatment doubly robust estimator in which $\mu(t,x)=\mathbb{E}[Y \mid T=t, X=x]$ is estimated using a Kernel Neural Network, and the conditional density $f_{T\mid X}(t \mid x)$ is estimated using a MultiGPS model. VCNet \citep{nie2021vcnet}, which jointly models outcomes and GPS; and VCNet-TR, which augments VCNet with targeted regularization. 

\textbf{Evaluation.} We use the integrated root mean squared error (IRMSE) as our evaluation metric
\[
\mathrm{IRMSE} = \int \left[ \frac{1}{S} \sum_{s=1}^{S} \left\{ \widehat{\varphi}_{s}(t) - \varphi(t) \right\}^{2} \right]^{1/2} \widehat{f}(t) \, \mathrm{d}t,
\]
where \( \widehat{f}(t) \) is the kernel density estimate of the treatment distribution and $s$ is the time of simulation replication. We use Monte Carlo methods to approximate the integral \citep{rubinstein2016simulation}.

\textbf{Results.} Table~\ref{tab:table1_final_mean} summarizes IRMSE across datasets, sample sizes, and covariate dimensions. WSENet-WTR achieves the best or near-best performance in nearly every configuration, with the largest gains at high dimension. Among weighting methods, DCOW consistently outperforms GPS, CBPS, and GBM, confirming that direct distributional balancing is more stable than density-ratio weighting as dimensionality grows. Among deep baselines, KernelNN-DR degrades sharply at high dimension, where conditional density estimation suffers from the curse of dimensionality, and VCNet-TR grows unstable on News as covariate count rises because its targeted regularization inverts an increasingly unreliable GPS estimate. On large-sample TCGA, WSENet and WSENet-WTR converge as plug-in bias vanishes, consistent with Theorem~1.

WSENet without WTR reveals an instructive failure mode. At $n=500$ with 25 covariates on IHDP its IRMSE deteriorates to 0.96 because DCOW decorrelates $X$ and $T$ while the reweighted sample concentrates mass where the outcome model is poorly constrained, amplifying plug-in bias in sparse regions. WSENet-WTR on the same setting collapses to 0.29, confirming that the EIF-based correction absorbs exactly this residual finite-sample bias. On TCGA, WSENet-WTR improves from 4.10e-3 to 2.10e-3 as covariates grow from 1000 to 4000 at $n=6000$, since DCOW needs no density model and the spline expansion acts on the scalar treatment independently of $p$; ACFR, by contrast, shows only modest gains. Together these results demonstrate the value of combining distributional balancing, spline-expanded treatment embeddings, and EIF-based correction, particularly in finite-sample and high-dimensional regimes.

The supplementary material further provides an ablation study isolating the contribution of each component (spline expansion, DCOW weighting, and EIF-based correction) and an analysis of the effect of network depth.

\section{Real-World Data Application}
We apply our method to estimate the effect of fine particulate matter (PM2.5) exposure on cardiovascular mortality rates (CMR) using a county-level U.S. dataset spanning 1990-2010 \citep{wyatt2020contribution}. The treatment variable is annual PM2.5 concentration ($\mu$g/m$^3$), and the outcome is the annual CMR, measured as deaths per 100K individuals. Covariates include socioeconomic and housing characteristics from U.S. Census data across 1990, 2000, and 2010, such as unemployment rates, income, educational attainment, housing conditions, and healthcare access. These variables provide a rich set of confounders for causal analysis. To align temporally with available covariates, we focus on treatment and outcome data from the year 2000.

\begin{figure}[t]
\centering
\includegraphics[width=0.48\columnwidth]{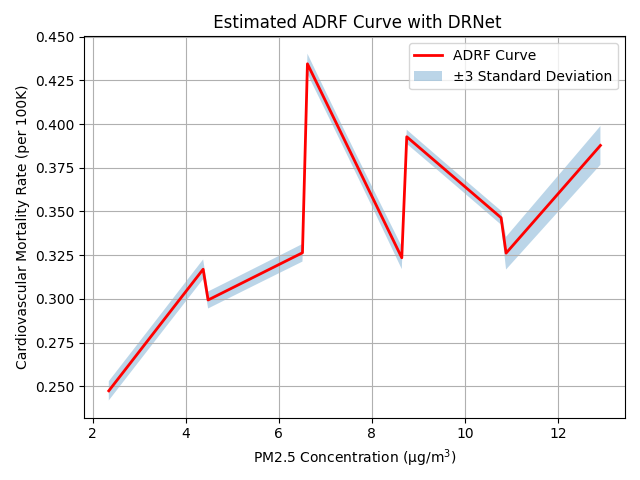}\hfill
\includegraphics[width=0.48\columnwidth]{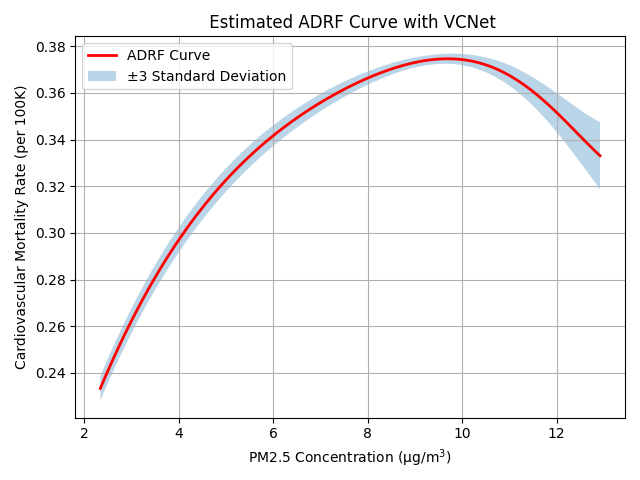}\\[2pt]
\includegraphics[width=0.48\columnwidth]{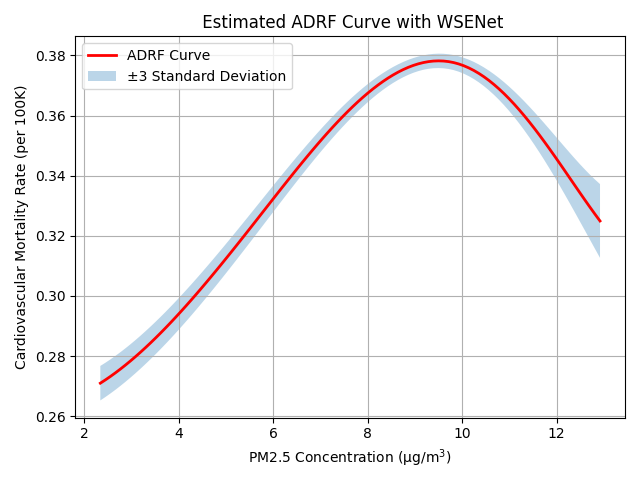}\hfill
\includegraphics[width=0.48\columnwidth]{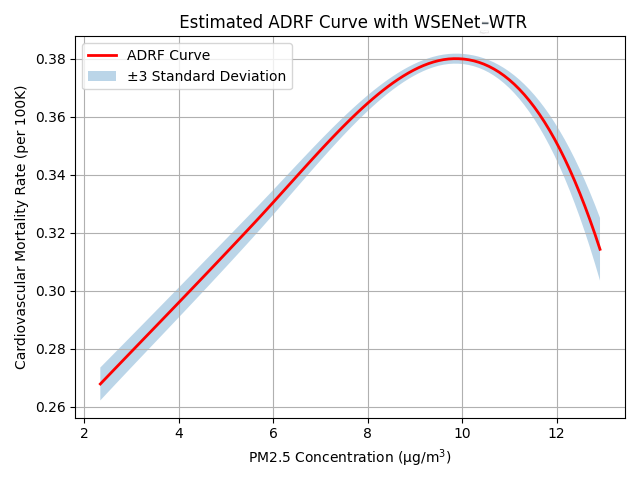}
\caption{ADRF curves by different deep learning methods.}
\label{fig:adrf}
\end{figure}

We then estimate the Average Dose-Response Function (ADRF) using four deep learning models. DRNet, VCNet, WSENet, and WSENet-WTR. The data is randomly split into training and testing sets with a 2:1 ratio. To avoid extrapolation beyond regions supported by data, we restrict the treatment range to lie within three standard deviations from its mean. The resulting ADRF estimates are shown in Figure~\ref{fig:adrf}. DRNet yields a highly variable and jagged ADRF curve, indicating instability likely due to discretization and head-splitting across treatment bins. VCNet shows smoother trends but with inflated variance in the tail regions, reflecting sensitivity to model regularization and limited overlap. WSENet, which incorporates distributional balancing weights, improves overall smoothness and variance control. WSENet-WTR achieves the most stable result and produces a smooth, well-regularized ADRF with narrow confidence bands throughout the treatment range. 

From the ADRF curves, we observe a non-monotonic relationship between PM2.5 concentration and cardiovascular mortality rates (CMR). CMR increases with PM2.5 exposure at lower concentrations, reaching a peak around 10~$\mu$g/m$^3$, beyond which it begins to decline gradually. This inverted-U pattern suggests a possible saturation effect, where the marginal harm of additional pollution diminishes in highly exposed regions. One possible explanation is that counties with higher pollution levels may also have more robust healthcare infrastructure or higher socioeconomic status, which could buffer the adverse health impacts of pollution. Alternatively, this trend may be partially driven by unmeasured confounders that vary regionally and influence both pollution exposure and health outcomes. These findings highlight the necessity of using flexible, bias-corrected estimators, such as WSENet-WTR, that can accommodate complex and nonlinear dose-response relationships. Traditional methods that assume monotonic or linear effects may fail to capture these nuanced patterns, leading to inaccurate or oversimplified conclusions in environmental health studies.

\section{Conclusion}
We introduced the WSENet, a novel framework for estimating the ADRF under continuous treatments. WSENet combines distributional balancing weights to reduce confounding bias, spline expansions to flexibly model treatment effects, and weighted targeted regularization for bias correction. Experiments on semi-synthetic and real-world data demonstrate its superior performance over existing kernel and deep learning-based ADRF estimators, especially in high-dimensional settings.

Our approach has several limitations that point to important directions for future work; we discuss them in detail in the supplementary material. Briefly, WSENet relies on the standard ignorability and (weak) positivity assumptions, which are typically unverifiable in observational studies and may fail locally under continuous treatments. Importantly, ignorability is an identification assumption and does not by itself guarantee that the observed data are empirically balanced; it is precisely this gap that motivates the explicit adjustment DCOW performs, and when the assumption is violated estimates can remain biased even when balance on measured covariates looks good. Promising remedies include proximal causal inference with proxy variables, sensitivity analysis via dependence perturbation of the balancing weights, falsification tests for unconfoundedness, and alternative estimands such as modified treatment policies under weak overlap. Finally, like many deep learning-based estimators, WSENet is largely a black box, and improving interpretability via feature attribution, counterfactual explanations, or interpretable surrogate models could make it more actionable for scientific and policy use.

\section*{Acknowledgements}
This was supported in part by the National Institutes of Health/National Institute of General Medical Sciences 1R01GM169395 (CP, GC), and the Patient-Centered Outcomes Research Institute (PCORI) Award ME-2024C1-37433 (GC). The statements in this work are solely the responsibility of the authors and do not necessarily represent the views of the Patient-Centered Outcomes Research Institute (PCORI), its Board of Governors, or the Methodology Committee.

\bibliographystyle{plainnat}
\bibliography{main}

\clearpage
\appendix

\begin{center}
  {\Large\bfseries Supplementary Material}
\end{center}

\section{Additional Experimental Results}
\subsection{Ablation Study}
To quantify the contribution of each design 
component, we conduct ablation experiments on IHDP and News. Table~\ref{tab:table2_final_ci} reports results for 
five variants. \textit{Weighted MLP} replaces the B-spline treatment 
expansion with naive concatenation of the raw treatment scalar and the 
covariate representation, isolating the contribution of structured 
treatment embedding. \textit{GPSNet} retains the spline expansion and 
network architecture of WSENet but substitutes DCOW with GPS-based 
inverse probability weighting, isolating the contribution of 
distributional balancing. \textit{GPSNet-WTR} adds weighted targeted 
regularization on top of GPSNet.

Weighted MLP degrades sharply relative to all spline-based variants, 
with IRMSE roughly doubling on IHDP and increasing by a factor of three 
or more on News, confirming that structured treatment expansion is 
essential for capturing nonlinear dose-response effects. Replacing DCOW 
with GPS-based weighting leaves IRMSE comparable on IHDP but introduces 
severe instability on News, where GPSNet-WTR reaches $0.530 \pm 0.32$ at 
$n=1000$ with 100 covariates: the curse of dimensionality inflates 
variance in the tails of the GPS estimate, inverse weighting places 
extreme emphasis on a small number of observations, and the WTR 
correction term inherits and amplifies this instability across 
replications. DCOW weights are bounded by construction and require no 
density inversion, which is why WSENet-WTR on the same configuration 
achieves $0.083 \pm 0.02$. WSENet-WTR achieves the lowest IRMSE in every 
configuration, with the margin over GPSNet-WTR widening as sample size 
decreases, consistent with WTR's role in correcting finite-sample bias 
that GPS-based weighting fails to absorb. Together, these results confirm 
that spline expansion, distributional balancing via DCOW, and EIF-based 
correction each contribute independently and that their combination is 
necessary for WSENet's full performance.

\begin{table*}[th]
  \centering
  \renewcommand{\arraystretch}{1.35}
   \caption{Ablation study: IRMSE with 95\% confidence intervals on IHDP 
  and News.}
  \resizebox{\textwidth}{!}{%
  \begin{tabular}{lcccccccccc}
    \toprule
    \textbf{Dataset} & \multicolumn{4}{c}{\textbf{IHDP}} & \multicolumn{6}{c}{\textbf{News}} \\
    \cmidrule(lr){1-1} \cmidrule(lr){2-5} \cmidrule(lr){6-11}
    \textbf{Num} & \multicolumn{2}{c}{\textbf{200}} & \multicolumn{2}{c}{\textbf{500}} & \multicolumn{3}{c}{\textbf{1000}} & \multicolumn{3}{c}{\textbf{2000}} \\
    \cmidrule(lr){1-1} \cmidrule(lr){2-3} \cmidrule(lr){4-5} \cmidrule(lr){6-8} \cmidrule(lr){9-11}
    \textbf{Covariates} & \textbf{10} & \textbf{25} & \textbf{10} & \textbf{25} & \textbf{100} & \textbf{300} & \textbf{500} & \textbf{100} & \textbf{300} & \textbf{500} \\
    \cmidrule(lr){1-1} \cmidrule(lr){2-5} \cmidrule(lr){6-11}
    Weighted MLP & 1.91$_{\scriptstyle \pm 0.04}$ & 1.92$_{\scriptstyle \pm 0.04}$ & 1.92$_{\scriptstyle \pm 0.04}$ & 1.90$_{\scriptstyle \pm 0.05}$ & 0.517$_{\scriptstyle \pm 0.04}$ & 0.492$_{\scriptstyle \pm 0.01}$ & 0.492$_{\scriptstyle \pm 0.01}$ & 0.516$_{\scriptstyle \pm 0.02}$ & 0.505$_{\scriptstyle \pm 0.02}$ & 0.494$_{\scriptstyle \pm 0.01}$ \\
    GPSNet       & 0.54$_{\scriptstyle \pm 0.05}$ & 0.40$_{\scriptstyle \pm 0.04}$ & 0.40$_{\scriptstyle \pm 0.04}$ & 0.92$_{\scriptstyle \pm 0.02}$ & 0.657$_{\scriptstyle \pm 0.42}$ & 0.108$_{\scriptstyle \pm 0.04}$ & 0.155$_{\scriptstyle \pm 0.07}$ & 0.119$_{\scriptstyle \pm 0.00}$ & 0.162$_{\scriptstyle \pm 0.06}$ & 0.098$_{\scriptstyle \pm 0.02}$ \\
    GPSNet-WTR    & 0.47$_{\scriptstyle \pm 0.03}$ & 0.36$_{\scriptstyle \pm 0.06}$ & 0.20$_{\scriptstyle \pm 0.01}$ & 0.26$_{\scriptstyle \pm 0.03}$ & 0.530$_{\scriptstyle \pm 0.32}$ & 0.118$_{\scriptstyle \pm 0.05}$ & 0.176$_{\scriptstyle \pm 0.06}$ & 0.209$_{\scriptstyle \pm 0.00}$ & 0.167$_{\scriptstyle \pm 0.07}$ & 0.139$_{\scriptstyle \pm 0.04}$ \\
    \cmidrule(lr){1-1} \cmidrule(lr){2-5} \cmidrule(lr){6-11}
    WSENet       & 0.49$_{\scriptstyle \pm 0.06}$ & 0.96$_{\scriptstyle \pm 0.12}$ & 0.31$_{\scriptstyle \pm 0.03}$ & 0.88$_{\scriptstyle \pm 0.06}$ & 0.160$_{\scriptstyle \pm 0.03}$ & 0.150$_{\scriptstyle \pm 0.01}$ & 0.090$_{\scriptstyle \pm 0.02}$ & 0.112$_{\scriptstyle \pm 0.02}$ & 0.135$_{\scriptstyle \pm 0.04}$ & 0.144$_{\scriptstyle \pm 0.02}$ \\
    WSENet-WTR   & \textbf{0.34}$_{\scriptstyle \pm 0.05}$ & \textbf{0.29}$_{\scriptstyle \pm 0.03}$ & \textbf{0.17}$_{\scriptstyle \pm 0.00}$ & \textbf{0.18}$_{\scriptstyle \pm 0.02}$ & \textbf{0.083}$_{\scriptstyle \pm 0.02}$ & \textbf{0.074}$_{\scriptstyle \pm 0.01}$ & \textbf{0.078}$_{\scriptstyle \pm 0.02}$ & \textbf{0.065}$_{\scriptstyle \pm 0.02}$ & \textbf{0.049}$_{\scriptstyle \pm 0.03}$ & \textbf{0.058}$_{\scriptstyle \pm 0.00}$ \\
    \bottomrule
  \end{tabular}%
  }
 
  \label{tab:table2_final_ci}
\end{table*}

\subsection{The Impact of Network Depth}
WSENet-WTR's advantage over WSENet 
could reflect either the benefit of EIF-based correction or simply 
insufficient model capacity in the 2-layer encoder. To disentangle these, 
we test encoder depths of 2, 3, and 4 layers on IHDP and News.

Table~\ref{tab:table3_depth_ci} shows that on IHDP, deeper encoders 
narrow the WSENet--WSENet-WTR gap, from 0.67 at 2 layers to 0.02 at 3 
layers, confirming that greater expressiveness absorbs some finite-sample 
bias. On the higher-dimensional News settings, however, increasing depth 
hurts: 3- and 4-layer WSENet-WTR consistently underperform the 2-layer 
variant at 300 and 500 covariates, suggesting that larger networks overfit 
when the covariate dimension is high relative to sample size. WTR's 
benefit is therefore not reducible to capacity alone; it provides a 
principled correction that remains reliable precisely where deeper 
encoders begin to overfit. Since established neural ADRF baselines use 
2--3 hidden layers, our main comparisons adopt depth-2 architectures for 
fairness.

\begin{table*}[th]
  \centering
  \renewcommand{\arraystretch}{1.35}
    \caption{Effect of network depth: IRMSE with 95\% confidence intervals 
  for WSENet and WSENet-WTR}
  \resizebox{\textwidth}{!}{%
  \begin{tabular}{lcccccccccc}
    \toprule
    \textbf{Dataset} & \multicolumn{4}{c}{\textbf{IHDP}} & \multicolumn{6}{c}{\textbf{News}} \\
    \cmidrule(lr){1-1} \cmidrule(lr){2-5} \cmidrule(lr){6-11}
    \textbf{Num} & \multicolumn{2}{c}{\textbf{200}} & \multicolumn{2}{c}{\textbf{500}} & \multicolumn{3}{c}{\textbf{1000}} & \multicolumn{3}{c}{\textbf{2000}} \\
    \cmidrule(lr){1-1} \cmidrule(lr){2-3} \cmidrule(lr){4-5} \cmidrule(lr){6-8} \cmidrule(lr){9-11}
    \textbf{Covariates} & \textbf{10} & \textbf{25} & \textbf{10} & \textbf{25} & \textbf{100} & \textbf{300} & \textbf{500} & \textbf{100} & \textbf{300} & \textbf{500} \\
    \cmidrule(lr){1-1} \cmidrule(lr){2-5} \cmidrule(lr){6-11}
    WSENet (2-layer)     & 0.49$_{\scriptstyle \pm 0.06}$ & 0.96$_{\scriptstyle \pm 0.12}$ & 0.31$_{\scriptstyle \pm 0.03}$ & 0.88$_{\scriptstyle \pm 0.06}$ & 0.160$_{\scriptstyle \pm 0.03}$ & 0.150$_{\scriptstyle \pm 0.01}$ & 0.090$_{\scriptstyle \pm 0.02}$ & 0.112$_{\scriptstyle \pm 0.02}$ & 0.135$_{\scriptstyle \pm 0.04}$ & 0.144$_{\scriptstyle \pm 0.02}$ \\
    WSENet-WTR (2-layer) & 0.34$_{\scriptstyle \pm 0.05}$ & 0.29$_{\scriptstyle \pm 0.03}$ & 0.17$_{\scriptstyle \pm 0.00}$ & 0.18$_{\scriptstyle \pm 0.02}$ & 0.083$_{\scriptstyle \pm 0.02}$ & 0.074$_{\scriptstyle \pm 0.01}$ & 0.078$_{\scriptstyle \pm 0.02}$ & 0.065$_{\scriptstyle \pm 0.02}$ & 0.049$_{\scriptstyle \pm 0.03}$ & 0.058$_{\scriptstyle \pm 0.00}$ \\
    WSENet (3-layer)     & 0.35$_{\scriptstyle \pm 0.06}$ & 0.28$_{\scriptstyle \pm 0.03}$ & 0.26$_{\scriptstyle \pm 0.00}$ & 0.23$_{\scriptstyle \pm 0.00}$ & 0.122$_{\scriptstyle \pm 0.02}$ & 0.099$_{\scriptstyle \pm 0.02}$ & 0.092$_{\scriptstyle \pm 0.02}$ & 0.086$_{\scriptstyle \pm 0.02}$ & 0.099$_{\scriptstyle \pm 0.03}$ & 0.092$_{\scriptstyle \pm 0.01}$ \\
    WSENet-WTR (3-layer) & 0.27$_{\scriptstyle \pm 0.04}$ & 0.26$_{\scriptstyle \pm 0.04}$ & 0.14$_{\scriptstyle \pm 0.00}$ & 0.21$_{\scriptstyle \pm 0.00}$ & 0.118$_{\scriptstyle \pm 0.02}$ & 0.092$_{\scriptstyle \pm 0.01}$ & 0.111$_{\scriptstyle \pm 0.01}$ & 0.080$_{\scriptstyle \pm 0.02}$ & 0.086$_{\scriptstyle \pm 0.00}$ & 0.098$_{\scriptstyle \pm 0.02}$ \\
    WSENet (4-layer)     & 0.24$_{\scriptstyle \pm 0.03}$ & 0.27$_{\scriptstyle \pm 0.03}$ & 0.18$_{\scriptstyle \pm 0.00}$ & 0.18$_{\scriptstyle \pm 0.01}$ & 0.111$_{\scriptstyle \pm 0.02}$ & 0.077$_{\scriptstyle \pm 0.02}$ & 0.089$_{\scriptstyle \pm 0.02}$ & 0.072$_{\scriptstyle \pm 0.05}$ & 0.074$_{\scriptstyle \pm 0.01}$ & 0.081$_{\scriptstyle \pm 0.00}$ \\
    WSENet-WTR (4-layer) & 0.20$_{\scriptstyle \pm 0.02}$ & 0.22$_{\scriptstyle \pm 0.02}$ & 0.12$_{\scriptstyle \pm 0.02}$ & 0.15$_{\scriptstyle \pm 0.00}$ & 0.103$_{\scriptstyle \pm 0.02}$ & 0.082$_{\scriptstyle \pm 0.01}$ & 0.105$_{\scriptstyle \pm 0.02}$ & 0.069$_{\scriptstyle \pm 0.03}$ & 0.070$_{\scriptstyle \pm 0.00}$ & 0.081$_{\scriptstyle \pm 0.01}$ \\
    \bottomrule
  \end{tabular}%
  }

  \label{tab:table3_depth_ci}
\end{table*}

\section{Covariate Balance Diagnostics for the Real-World Application}
\begin{figure}[t]
    \centering
    \includegraphics[width=0.75\textwidth]{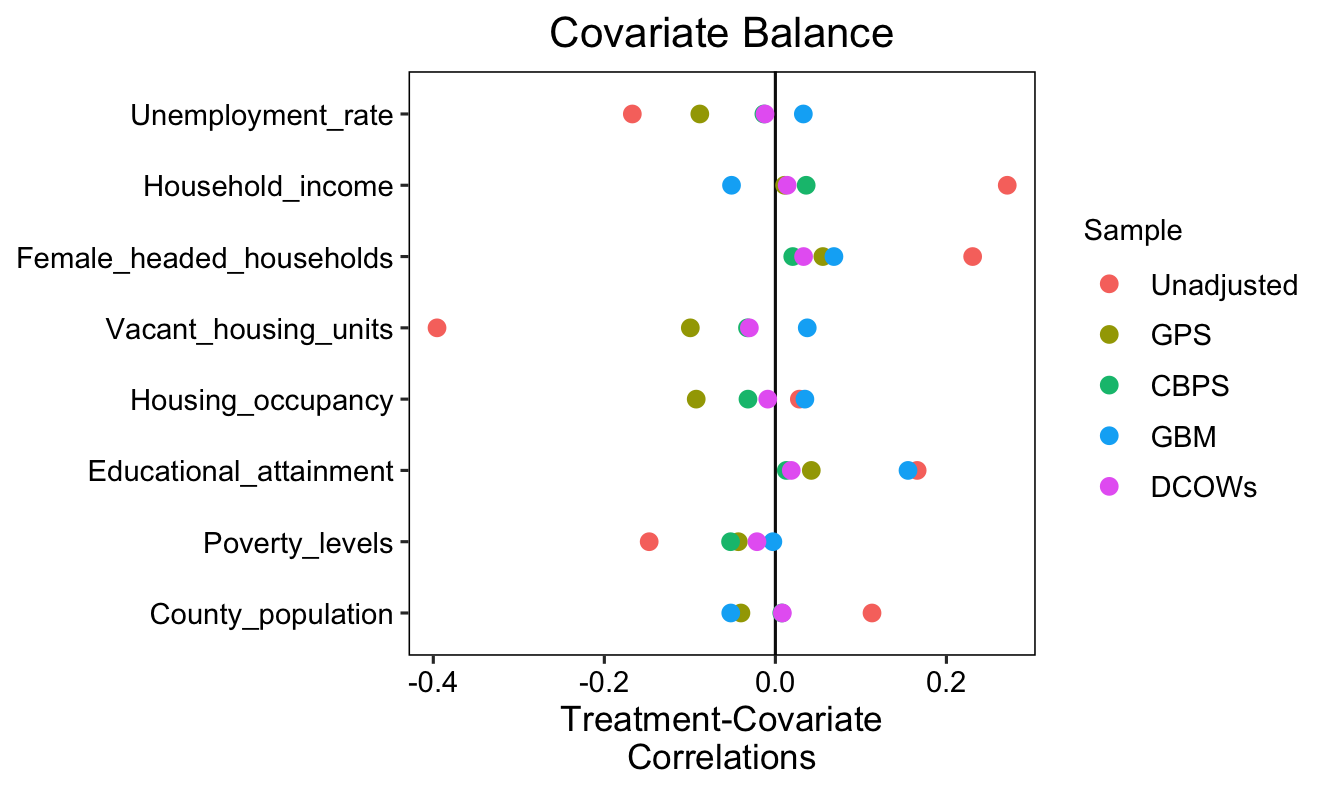}
    \caption{Love plot for different balancing weights}
    \label{fig:loveplot}
\end{figure}

We evaluate the covariate balance of the PM2.5 dataset before and after weighting using standardized mean differences, implemented via the \texttt{cobalt} R package \citep{greifer2020covariate}. As shown in Figure~\ref{fig:loveplot}, the unweighted data exhibit substantial imbalance across multiple covariates. Both CBPS and DCOW substantially improve balance, with DCOW achieving the greatest reduction in standardized mean differences across most covariates. These results highlight the presence of severe confounding in the unweighted data and demonstrate that DCOW provides the most effective confounding adjustment, further motivating its use in the ADRF analysis of the main paper.

\section{Limitations and Future Work}
Our approach has several limitations that point to important directions for future work. First, WSENet relies on the standard ignorability (unconfoundedness) assumption for continuous treatments, namely that all common causes of $T$ and $Y$ are observed and appropriately adjusted for---a strong requirement that is typically unverifiable in observational studies \citep{hernan2004structural}. Importantly, ignorability is an identification assumption and does not guarantee that the observed data are empirically balanced; rather, it motivates the need for explicit adjustment. When ignorability is violated, the model remains sensitive to potential unmeasured confounding, and causal estimates can be biased even if the outcome model is highly flexible and the weighting achieves good balance on measured covariates. To address this limitation, which is shared by nearly all observational methods, future work can pursue two main paths. One direction is to integrate frameworks that relax ignorability, such as proximal causal inference using proxy variables \citep{miao2018, TT2024}, or to develop hybrid architectures that incorporate instrumental-variable structures when valid instruments are available. Another critical direction is conducting sensitivity analysis to assess how violations of unconfoundedness might affect the estimated ADRF. While most existing sensitivity frameworks \citep{bonvini2022sensitivity} are developed for binary treatments, they highlight structural ideas that can motivate continuous extensions. For weighting-based approaches like ours, an alternative strategy is to assess sensitivity via dependence perturbation, wherein one systematically relaxes the treatment-covariate independence induced by balancing weights (e.g., via DCOW) to simulate the impact of residual confounding. Additionally, recent work \citep{karlsson2025falsification} introduces a falsification test for unconfoundedness under continuous exposures. We plan to incorporate such tools to strengthen the robustness and transparency of ADRF estimation.

Second, WSENet, like most ADRF estimators, assumes (weak) positivity/overlap. In continuous-treatment settings, overlap can fail locally and is difficult to diagnose. When overlap is weak, a principled remedy is often to target alternative estimands that do not require strict positivity, such as modified treatment policies or shift/incremental interventions \citep{haneuse2013estimation, schindl2024incremental}, or to adapt the estimand to feasible treatment regions \citep{bao2025addressing, zhang2024nonparametric}. Finally, like many deep learning-based estimators, WSENet is largely a black box, which limits interpretability---for example, how individual covariates contribute to the estimated ADRF or whether learned representations reflect meaningful effect-modifying structure. Improving transparency via feature attribution, counterfactual explanations, or interpretable surrogate models could make WSENet more actionable for scientific and policy use. Together, these directions highlight opportunities to extend WSENet into a more interpretable, robust, and reliable framework for continuous-treatment causal inference.

\section{Independence Weights Details}
\subsection{Introduction and Objective}
To achieve robust deconfounding and induce independence between covariates $\mathbf{X} \in \mathbb{R}^p$ and a treatment variable $T \in \mathbb{R}$, we employ Distance Covariate Optimal Weights (DCOWs) \citep{huling2024independence}. DCOWs provide a non-parametric, assumption-free method to estimate weights $\mathbf{w} = (w_1, \ldots, w_n)^\intercal$ for a sample of size $n$. The fundamental goal is to reweight the observed data such that the joint empirical cumulative distribution function (CDF) of covariates and treatment in the weighted sample approximates the product of their respective marginal empirical CDFs from the original unweighted sample. That is, we seek weights $\mathbf{w}$ such that
$F_{\mathbf{X}, T, \mathbf{w}}^n(\mathbf{x}, t) \approx F_{\mathbf{X}}^n(\mathbf{x}) F_T^n(t)$.

Here, the empirical CDFs are defined by
\setlist[itemize]{left=0pt}
\begin{itemize}
\item   $F_{\mathbf{X}}^n(\mathbf{x}) = n^{-1} \sum_{i=1}^n I(\mathbf{X}_i \leq \mathbf{x})$ is the empirical CDF of the covariates $\{\mathbf{X}_i\}_{i=1}^n$ and similarly, $F_T^n(t) = n^{-1} \sum_{i=1}^n I(T_i \leq t)$.
\item   $F_{\mathbf{X}, T, \mathbf{w}}^n(\mathbf{x}, t) = n^{-1} \sum_{i=1}^n w_i I(\mathbf{X}_i \leq \mathbf{x}, T_i \leq t)$ is the weighted empirical CDF of the joint distribution of covariates and treatment, using weights $\mathbf{w}$ constrained such that $\sum_{i=1}^n w_i = n$ and $w_i \geq 0$ for all $i$.
\end{itemize}
Achieving this approximate independence, or "distributional decorrelation," is key to mitigating confounding biases.

\subsection{The DCOW Objective Function $\mathcal{D}(\mathbf{w})$}
The DCOW method estimates the optimal weights $\mathbf{w}$ by minimizing the objective function $\mathcal{D}(\mathbf{w})$
$$ \mathcal{D}(\mathbf{w}) = \mathcal{V}_{n,\mathbf{w}}^2(\mathbf{X}, T) + \mathcal{E}(F_{\mathbf{X},\mathbf{w}}^n, F_{\mathbf{X}}^n) + \mathcal{E}(F_{T,\mathbf{w}}^n, F_T^n). $$
This objective function is composed of three critical terms
\begin{itemize}
    \item   $\mathcal{V}_{n,\mathbf{w}}^2(\mathbf{X}, T)$: A weighted distance covariance term that quantifies the joint dependence between $\mathbf{X}$ and $T$ after weighting.
    \item   $\mathcal{E}(F_{\mathbf{X},\mathbf{w}}^n, F_{\mathbf{X}}^n)$: An energy distance term that measures the discrepancy between the marginal distribution of covariates in the weighted sample ($F_{\mathbf{X},\mathbf{w}}^n$) and the original empirical marginal distribution ($F_{\mathbf{X}}^n$). $\mathcal{E}(F_{T,\mathbf{w}}^n, F_T^n)$ defines similarly.
\end{itemize}
Minimizing $\mathcal{D}(\mathbf{w})$ aims to find weights that render $\mathbf{X}$ and $T$ approximately independent while ensuring that the marginal distributions of $\mathbf{X}$ and $T$ in the weighted sample remain faithful to their original empirical distributions.

\subsubsection{Weighted Distance Covariance Term $\mathcal{V}_{n,\mathbf{w}}^2(\mathbf{X}, T)$}
This term measures the dependence between $\mathbf{X}$ and $T$ in the weighted sample. It is adapted from the distance covariance concept \citep{szekely2007measuring} and is defined using empirical characteristic functions
\begin{align*}
\mathcal{V}_{n, \mathbf{w}}^2(\mathbf{X}, T) = \int \Big| &\varphi_{\mathbf{X}, T, \mathbf{w}}^n(\mathbf{m}, \nu) - \varphi_{\mathbf{X}, \mathbf{w}}^n(\mathbf{m}) \varphi_{T, \mathbf{w}}^n(\nu) \\ &+ \big(\varphi_{\mathbf{X}, \mathbf{w}}^n(\mathbf{m}) - \varphi_{\mathbf{X}}^n(\mathbf{m})\big) \big(\varphi_{T, \mathbf{w}}^n(\nu) - \varphi_T^n(\nu)\big) \Big|^2 \omega(\mathbf{m}, \nu) \, d\mathbf{m} \, d\nu.
\end{align*}
The components are
\begin{itemize}
    \item   $\varphi_{\mathbf{X}, T, \mathbf{w}}^n(\mathbf{m}, \nu) = \frac{1}{n} \sum_{j=1}^n w_j \exp\left\{i \mathbf{m}^\intercal \mathbf{X}_j + i \nu T_j\right\}$: The empirical characteristic function of the joint distribution $(\mathbf{X}, T)$ using weights $\mathbf{w}$.
    \item   $\varphi_{\mathbf{X}, \mathbf{w}}^n(\mathbf{m}) = \frac{1}{n} \sum_{j=1}^n w_j \exp\left\{i \mathbf{m}^\intercal \mathbf{X}_j\right\}$: The empirical characteristic function of $\mathbf{X}$ using weights $\mathbf{w}$ ($F_{\mathbf{X},\mathbf{w}}^n$ is its corresponding CDF) and similarly, $\varphi_{T, \mathbf{w}}^n(\nu) = \frac{1}{n} \sum_{j=1}^n w_j \exp\left\{i \nu T_j\right\}$.
    \item   $\varphi_{\mathbf{X}}^n(\mathbf{m}) = \frac{1}{n} \sum_{j=1}^n \exp\left\{i \mathbf{m}^\intercal \mathbf{X}_j\right\}$: The empirical characteristic function of the original covariate distribution and similarly, $\varphi_T^n(\nu) = \frac{1}{n} \sum_{j=1}^n \exp\left\{i \nu T_j\right\}$.
    \item   $\omega(\mathbf{m}, \nu) = (c_p c_1 \|\mathbf{m}\|_2^{1+p} |\nu|^2)^{-1}$ is a weighting function, where $p$ is the dimension of $\mathbf{X}$, $c_k = \frac{\pi^{(1+k)/2}}{\Gamma((1+k)/2)}$, and $\Gamma(\cdot)$ is the gamma function.
\end{itemize}

\subsubsection{Marginal Distribution Preservation Terms: Energy Distances $\mathcal{E}$}
The goal of achieving $F_{\mathbf{X}, T, \mathbf{w}}^n = F_{\mathbf{X}}^n F_T^n$ requires not only that the weighted joint distribution factors into its weighted marginals (addressed by the first part of $\mathcal{V}_{n, \mathbf{w}}^2$) but also that these weighted marginals are close to the original unweighted marginals. The energy distance terms enforce this fidelity.

The second term, $\mathcal{E}(F_{\mathbf{X}, \mathbf{w}}^n, F_{\mathbf{X}}^n)$, measures the energy distance \citep{huling2024energy} between the weighted empirical CDF of covariates $F_{\mathbf{X}, \mathbf{w}}^n$ and the original empirical CDF $F_{\mathbf{X}}^n$
$$ \mathcal{E}(F_{\mathbf{X}, \mathbf{w}}^n, F_{\mathbf{X}}^n) = \int_{\mathbb{R}^p} |\varphi_{\mathbf{X}}^n(\mathbf{m}) - \varphi_{\mathbf{X}, \mathbf{w}}^n(\mathbf{m})|^2 \omega(\mathbf{m}) \, d\mathbf{m}, $$
where $\omega(\mathbf{m}) = \frac{1}{c_p \|\mathbf{m}\|_2^{1+p}}$ and $c_p = \frac{\pi^{(1+p)/2}}{\Gamma((1+p)/2)}$.

The third term, $\mathcal{E}(F_{T, \mathbf{w}}^n, F_T^n)$, similarly measures the energy distance between the weighted empirical CDF of the treatment $F_{T, \mathbf{w}}^n$ and its original empirical CDF $F_T^n$
$$ \mathcal{E}(F_{T, \mathbf{w}}^n, F_T^n) = \int_{\mathbb{R}} |\varphi_T^n(\nu) - \varphi_{T, \mathbf{w}}^n(\nu)|^2 \omega(\nu) \, d\nu, $$
where $\omega(\nu) = \frac{1}{c_1 |\nu|^{1+1}} = \frac{1}{\pi \nu^2}$ (since $T$ is typically univariate).

Minimizing these energy distances ensures that $F_{\mathbf{X}, \mathbf{w}}^n \approx F_{\mathbf{X}}^n$ and $F_{T, \mathbf{w}}^n \approx F_T^n$. When these conditions hold, minimizing $\mathcal{V}_{n, \mathbf{w}}^2(\mathbf{X}, T)$ effectively seeks $F_{\mathbf{X}, T, \mathbf{w}}^n(\mathbf{x},t) \approx F_{\mathbf{X},\mathbf{w}}^n(\mathbf{x})F_{T,\mathbf{w}}^n(t) \approx F_{\mathbf{X}}^n(\mathbf{x})F_T^n(t)$.

\subsection{Euclidean Forms for Computation}
The terms in $\mathcal{D}(\mathbf{w})$ involving characteristic functions can be expressed using Euclidean distances between sample observations, which facilitates computation \citep{huling2024energy}.

\paragraph{For $\mathcal{V}_{n,\mathbf{w}}^2(\mathbf{X}, T)$:}
The Euclidean form for $\mathcal{V}_{n,\mathbf{w}}^2(\mathbf{X}, T)$ is
$$ \mathcal{V}_{n,\mathbf{w}}^2(\mathbf{X}, T) = \frac{1}{n^2} \sum_{k, \ell = 1}^n w_k w_\ell C_{k \ell} D_{k \ell}, $$
where
$c_{k \ell} = \|\mathbf{X}_k - \mathbf{X}_\ell\|_2$, $\bar{c}_{k \cdot} = \frac{1}{n} \sum_{\ell = 1}^n c_{k \ell}$, $\bar{c}_{\cdot \ell} = \frac{1}{n} \sum_{k = 1}^n c_{k \ell}$, $\bar{c}_{\cdot \cdot} = \frac{1}{n^2} \sum_{k, \ell = 1}^n c_{k \ell}$, and $C_{k \ell} = c_{k \ell} - \bar{c}_{k \cdot} - \bar{c}_{\cdot \ell} + \bar{c}_{\cdot \cdot}$.
Similarly, for treatment $T$,
$d_{k \ell} = |T_k - T_\ell|$, $\bar{d}_{k \cdot} = \frac{1}{n} \sum_{\ell = 1}^n d_{k \ell}$, $\bar{d}_{\cdot \ell} = \frac{1}{n} \sum_{k = 1}^n d_{k \ell}$, $\bar{d}_{\cdot \cdot} = \frac{1}{n^2} \sum_{k, \ell = 1}^n d_{k \ell}$, and $D_{k \ell} = d_{k \ell} - \bar{d}_{k \cdot} - \bar{d}_{\cdot \ell} + \bar{d}_{\cdot \cdot}$.

\paragraph{For $\mathcal{E}(F_{\mathbf{X}, \mathbf{w}}^n, F_\mathbf{X}^n)$:}
The energy distance between $F_{\mathbf{X}, \mathbf{w}}^n$ and $F_\mathbf{X}^n$ (with Euclidean norm) is
$$ \mathcal{E}(F_{\mathbf{X}, \mathbf{w}}^n, F_\mathbf{X}^n) = \frac{2}{n^2} \sum_{i=1}^n w_i \sum_{j=1}^n \|\mathbf{X}_i - \mathbf{X}_j\|_2 - \frac{1}{n^2} \sum_{i=1}^n \sum_{j=1}^n w_i w_j \|\mathbf{X}_i - \mathbf{X}_j\|_2 - \frac{1}{n^2} \sum_{i=1}^n \sum_{j=1}^n \|\mathbf{X}_i - \mathbf{X}_j\|_2. $$

\paragraph{For $\mathcal{E}(F_{T, \mathbf{w}}^n, F_T^n)$:}
Similarly, for the treatment variable $T$
$$ \mathcal{E}(F_{T, \mathbf{w}}^n, F_T^n) = \frac{2}{n^2} \sum_{i=1}^n w_i \sum_{j=1}^n |T_i - T_j| - \frac{1}{n^2} \sum_{i=1}^n \sum_{j=1}^n w_i w_j |T_i - T_j| - \frac{1}{n^2} \sum_{i=1}^n \sum_{j=1}^n |T_i - T_j|. $$
These Euclidean forms allow $\mathcal{D}(\mathbf{w})$ to be expressed as a function of inter-sample distances and weights.

\section{Details of Efficient Influence Function (EIF)}
\textbf{Proposition B.1:} The efficient influence function of integrated ADRF $\psi=\int\int\mu(\textbf{x}, t)f(\textbf{x})f(t)dtdx$ is 
\begin{align}
 \frac{(Y - \mu(\boldsymbol{X}, T)) \, f(\boldsymbol{X}) \, f(T)}{f(\boldsymbol{X}, T)} \nonumber + \int\mu(\textbf{x}, T)d\mathbb{F}(\textbf{x}) - \psi \nonumber  + \int \left\{\mu(\textbf{X}, t) - \int \mu(\textbf{x}, t) \, d\mathbb{F}(\textbf{x}) \right\} f(t) \, dt 
. \nonumber
\end{align}

The empirical form of the efficient influence function is doubly robust: if either \( \hat{w} = w \) or \( \hat{\mu} = \mu \), then
\[
\hat{\xi}(\textbf{X}, Y, T, \hat{w}, \hat{\mu}) = \left(Y - \hat{\mu}(\textbf{X}, T)\right)\hat{w}(\textbf{X}, T) + \frac{1}{n} \sum_{j=1}^{n} \hat{\mu}(\textbf{X}_j, T)
\]
is also doubly robust \citep{kennedy2017non, van2011targeted}. We have
\[
\mathbf{E}(\hat{\xi}(\textbf{X}, Y, T, \hat{w}, \hat{\mu}) \mid T = t) = E_X[\mu(\textbf{X}, t)] = E(Y(t)).
\]

\textbf{Lemma 1: Doubly Robust Property of the Estimator }

Let $\mu(X,T) = E[Y \mid X,T]$ be the true outcome model and $w(X,T) = \frac{f(T)f(X)}{f(X,T)}$ be the true importance weight. The estimator $\hat{\xi}$ is defined as
\[ \hat{\xi}(X, Y, T, \hat{w}, \hat{\mu}) = \left(Y - \hat{\mu}(X, T)\right)\hat{w}(X, T) + \frac{1}{n} \sum_{j=1}^{n} \hat{\mu}(X_j, T). \]
Then, $E[\hat{\xi}(X, Y, T, \hat{w}, \hat{\mu}) \mid T = t] = E_X[\mu(X, t)]$ if $\hat{\mu}(X,T) = \mu(X,T)$ (outcome model is correct), or if $\hat{w}(X,T) = w(X,T)$ (weight model is correct).
(Note: $E_X[g(X,t)]$ denotes the expectation of $g(X,t)$ where $X \sim f(X)$.)

\begin{proof}

Given $T=t$, we know that $E\left[\frac{1}{n} \sum_{j=1}^{n} \hat{\mu}(X_j, t) \mid T=t\right] = E_X[\hat{\mu}(X,t)]$.
The conditional expectation of $\hat{\xi}$ is
\begin{align*} E[\hat{\xi} \mid T=t] &= E_{X|t}\left[\left(E[Y \mid X,t] - \hat{\mu}(X,t)\right)\hat{w}(X,t)\right] + E_X[\hat{\mu}(X,t)] \\ &= E_{X|t}\left[\left(\mu(X,t) - \hat{\mu}(X,t)\right)\hat{w}(X,t)\right] + E_X[\hat{\mu}(X,t)]. \end{align*}
Let $B = E_{X|t}\left[\left(\mu(X,t) - \hat{\mu}(X,t)\right)\hat{w}(X,t)\right]$.
Thus, $E[\hat{\xi} \mid T=t] = B + E_X[\hat{\mu}(X,t)]$. We need to show this equals $E_X[\mu(X,t)]$.

If $\hat{\mu}(X,t) = \mu(X,t)$ (outcome model correct),
Then $\mu(X,t) - \hat{\mu}(X,t) = 0$, so $B = 0$.
$E[\hat{\xi} \mid T=t] = 0 + E_X[\hat{\mu}(X,t)] = E_X[\mu(X,t)]$. The result holds.

If $\hat{w}(X,t) = w(X,t)$ (weight model correct),
The true weight $w(X,t) = \frac{f(X)}{f(X|T=t)}$.
\begin{align*} B &= E_{X|t}\left[\left(\mu(X,t) - \hat{\mu}(X,t)\right)w(X,t)\right] \\ &= \int \left(\mu(x,t) - \hat{\mu}(x,t)\right) \frac{f(x)}{f(x|t)} f(x|t) dx \\ &= \int \left(\mu(x,t) - \hat{\mu}(x,t)\right) f(x) dx \\ &= E_X[\mu(X,t)] - E_X[\hat{\mu}(X,t)]. \end{align*}
Substituting this into $E[\hat{\xi} \mid T=t] = B + E_X[\hat{\mu}(X,t)]$
\begin{align*} E[\hat{\xi} \mid T=t] &= \left(E_X[\mu(X,t)] - E_X[\hat{\mu}(X,t)]\right) + E_X[\hat{\mu}(X,t)] \\ &= E_X[\mu(X,t)]. \end{align*}
The result holds.

Thus, the estimator is doubly robust.
\end{proof}

\textbf{Proof of Proposition B.1}

\begin{proof}
Let $P_{\epsilon}$ be a one-dimensional parametric submodel such that $P_0 = P$. $f(Z;\epsilon)$ denotes the density function of the parametric submodel $P_{\epsilon}$. The score function at $\epsilon=0$ for an observation $Z=(X,T,Y)$ is $l'_{\epsilon}(Z;0) = \frac{\partial}{\partial\epsilon} \log f(Z;\epsilon)|_{\epsilon=0}$.
To show that $\phi(Z)$ is the efficient influence function, we need to check that it is mean-zero ($E[\phi(Z)]=0$), has finite variance ($E[\phi(Z)^2] < \infty$), and satisfies the pathwise differentiability condition
\begin{equation}
\frac{\partial}{\partial\epsilon}\psi(P_{\epsilon})|_{\epsilon=0} = E[\phi(Z)l_{\epsilon}^{\prime}(Z;0)].
\end{equation}
Let the influence function be
$$ \phi(Z) = \frac{(Y - \mu(X, T))w}{f(T \mid X)} + (\theta(T) - \psi) + \int (\mu(X,t) - \theta(t))w\,dt $$
where $w$ denotes a weighting function, which is $w(T)$ in the first term and $w(t)$ (or $w(s)$) in the integral. For the derivative check, we set $w(\cdot)$ to be the marginal density $f(\cdot;0)$.

\noindent\textbf{Checking mean zero:}
We want to show $E[\phi(Z)] = 0$.
\begin{align*} E[\phi(Z)] = E\left[\frac{(Y-\mu(X,T))w}{f(T \mid X)}\right] &+ E[\theta(T) - \psi] + E\left[\int (\mu(X,t)-\theta(t))w\,dt\right]. \end{align*}
The first term
\begin{align*} E\left[\frac{(Y-\mu(X,T))w}{f(T \mid X)}\right] &= E_{X,T}\left[ \frac{w}{f(T \mid X)} E[Y-\mu(X,T) \mid X,T] \right] = 0. \end{align*}
The third term
\begin{align*} E\left[\int (\mu(X,t)-\theta(t))w\,dt\right] &= \int (E_{X}[\mu(X,t)]-\theta(t))w\,dt = \int (\theta(t)-\theta(t))w\,dt = 0. \end{align*}
For the second term: $E[\theta(T) - \psi] = E[\theta(T)] - \psi = 0$ by definition of $\psi$.
Therefore, $E[\phi(Z)] = 0$.

\noindent\textbf{Checking finite variance:}
This follows from standard assumptions: positivity of $f(T|X)$ and finite variance of $Y$, $\mu(X,T)$, and $\theta(T)$.

\noindent\textbf{Checking pathwise differentiability:}
We decompose the joint score $l_{\epsilon}^{\prime}(Z;0)$ into orthogonal components
\[
l_{\epsilon}^{\prime}(Z;0) = l_{\epsilon}^{\prime}(Y \mid X,T;0) + l_{\epsilon}^{\prime}(T \mid X;0) + l_{\epsilon}^{\prime}(X;0).
\]
Let $\phi(Z) = \phi_1(Z) + \phi_2(Z) + \phi_3(Z)$, where $\phi_1$ is the weighted residual term, $\phi_2 = \theta(T) - \Psi$, and $\phi_3 = \int (\mu(X,s) - \theta(s)) f(s;0) ds$.

\subsection*{Left-Hand Side (LHS)}
The target parameter is $\Psi(P_{\epsilon}) = \iint \mu_\epsilon(x,t) f_\epsilon(x) f_\epsilon(t) dx dt$. Note that the marginal density of $T$ is induced by $f_\epsilon(t) = \int f_\epsilon(t|x)f_\epsilon(x)dx$.
Differentiating with respect to $\epsilon$ at $\epsilon=0$ yields three terms corresponding to the changes in $\mu$, $f_X$, and $f_T$
\begin{align*}
\frac{\partial}{\partial\epsilon}\Psi(P_{\epsilon})|_{\epsilon=0} &= \iint \mu_{\epsilon}'(x,t) f(x) f(t) dx dt \quad (\text{Term } L_1) \\
&\quad + \iint \mu(x,t) f_{\epsilon}'(x) f(t) dx dt \quad (\text{Term } L_2) \\
&\quad + \int \theta(t) f_{\epsilon}'(t) dt \quad (\text{Term } L_3).
\end{align*}
We now verify that $E[\phi(Z)l_{\epsilon}^{\prime}(Z;0)]$ recovers these three terms.

\subsection*{Right-Hand Side (RHS)}

\medskip
\noindent\textbf{Term involving $l_{\epsilon}^{\prime}(Y \mid X, T;0)$:}
Only $\phi_1(Z)$ depends on $Y$.
\begin{align*}
E[\phi(Z) l_{\epsilon}^{\prime}(Y \mid X, T)] &= E[\phi_1(Z) l_{\epsilon}^{\prime}(Y \mid X, T)] \\
&= E_{X,T}\left[ \frac{f(T)}{f(T \mid X)} E_Y[(Y-\mu(X,T)) l_{\epsilon}^{\prime}(Y \mid X, T) \mid X,T] \right].
\end{align*}
Using the identity $E_Y[(Y-\mu)l'] = \frac{\partial}{\partial \epsilon} E[Y|X,T] = \mu_{\epsilon}'(X,T)$, this becomes
\[
\iint \frac{f(t)}{f(t \mid x)} \mu_{\epsilon}'(x,t) f(t \mid x) f(x) dx dt = \iint \mu_{\epsilon}'(x,t) f(x) f(t) dx dt = L_1.
\]

\medskip
\noindent\textbf{Term involving $l_{\epsilon}^{\prime}(T \mid X;0)$:}
$\phi_1$ and $\phi_3$ are orthogonal to this score. Only $\phi_2 = \theta(T) - \Psi$ contributes.
\begin{align*}
E[\phi(Z) l_{\epsilon}^{\prime}(T \mid X)] &= E[(\theta(T) - \Psi) l_{\epsilon}^{\prime}(T \mid X)] \\
&= \iint \theta(t) l_{\epsilon}^{\prime}(t \mid x) f(t \mid x) f(x) dx dt \quad (\text{since } E[l'|X]=0 \implies \Psi \text{ term vanishes}) \\
&= \int \theta(t) \left( \int f_{\epsilon}^{\prime}(t \mid x) f(x) dx \right) dt. \quad (\text{Part A of } L_3)
\end{align*}

\medskip
\noindent\textbf{Term involving $l_{\epsilon}^{\prime}(X;0)$:}
This score captures the perturbation of the marginal distribution $f(x)$. Both $\phi_2$ (via dependency between $T$ and $X$) and $\phi_3$ contribute.
\begin{align*}
E[\phi(Z) l_{\epsilon}^{\prime}(X)] &= E[\phi_2(Z) l_{\epsilon}^{\prime}(X)] + E[\phi_3(Z) l_{\epsilon}^{\prime}(X)].
\end{align*}
For $\phi_2$
\begin{align*}
E[(\theta(T) - \Psi) l_{\epsilon}^{\prime}(X)] &= \iint \theta(t) f(t \mid x) l_{\epsilon}^{\prime}(x) f(x) dx dt \quad (\Psi \text{ term vanishes since } E[l'(X)]=0) \\
&= \int \theta(t) \left( \int f(t \mid x) f_{\epsilon}^{\prime}(x) dx \right) dt. \quad (\text{Part B of } L_3)
\end{align*}
Combining Part A and Part B recovers the full derivative of the marginal $f_T(t)$
\[
\text{Part A} + \text{Part B} = \int \theta(t) \left( \int (f_{\epsilon}^{\prime}(t \mid x)f(x) + f(t \mid x)f_{\epsilon}^{\prime}(x)) dx \right) dt = \int \theta(t) f_{\epsilon}^{\prime}(t) dt = L_3.
\]
For $\phi_3$
\begin{align*}
E[\phi_3(Z) l_{\epsilon}^{\prime}(X)] &= \int \left( \int (\mu(x,t) - \theta(t)) f(t) dt \right) f_{\epsilon}^{\prime}(x) dx \\
&= \iint \mu(x,t) f(t) f_{\epsilon}^{\prime}(x) dt dx - \int \theta(t) f(t) dt \underbrace{\int f_{\epsilon}^{\prime}(x) dx}_{0} \\
&= \iint \mu(x,t) f_{\epsilon}^{\prime}(x) f(t) dx dt = L_2.
\end{align*}

\medskip
Summing the results
\[ E[\phi(Z) l_{\epsilon}^{\prime}(Z)] = L_1 + L_3 + L_2 = \text{LHS}. \]
The condition is satisfied.
\end{proof}

\section{Proof of Theorem 1}

\textbf{Definitions}

To facilitate the proof of Theorem 1, we introduce several key definitions and notations.

Let \( \tilde{\epsilon}(t) \) be defined as the expectation of the residual term adjusted by the weighting function, conditioned on \( T = t \), and normalized by the expectation of the squared weighting function
\[
\tilde{\epsilon}(t) = \frac{\mathbb{E}\left[ (Y - \hat{\mu}_n(X, T)) \hat{\omega}_n \mid T = t \right]}{\mathbb{E}\left[\hat{\omega}_n^2 \mid T = t\right]}.
\]
The estimator \( \hat{\epsilon}(t) \) is represented as a linear combination of \( K_n \) B-spline basis functions \( \phi_k(t) \).  \( \hat{\epsilon}(t) = \sum_{k=1}^{K_n} \hat{\alpha}_k \phi_k(t) \), where \( \hat{\alpha} = (\hat{\alpha}_1, \dots, \hat{\alpha}_{K_n})^T \) is the vector of coefficients. We define the vector of these basis functions evaluated at \( t \) as \( \phi^{K_n}(t) = (N_{1,d}(t), \dots, N_{K_n,d}(t))^T \in \mathbb{R}^{K_n} \). The design matrix \( \Phi_n \in \mathbb{R}^{K_n \times n} \) is constructed by evaluating these basis vectors at each observed time point \( t_i \), i.e., \( \Phi_n = (\phi^{K_n}(t_1), \dots, \phi^{K_n}(t_n)) \), so that the \( (k,i) \)-th element of \( \Phi_n \) is \( \phi_k(t_i) \).

The weighting matrix \( \Lambda_n \) is the diagonal matrix whose entries are the inverses of the weighting functions evaluated at each observation
\[
\Lambda_n = \text{diag}\left( \hat{\omega}_1(t_1, x_1)^{-1}, \dots, \hat{\omega}_n(t_n, x_n)^{-1} \right).
\]
Similarly, \( \overset{\circ}{\Lambda}_n \) is the diagonal matrix whose entries are the inverses of the expected squared weighting function conditioned on \( T = t \)
\[
\overset{\circ}{\Lambda}_n = \text{diag}\left( \mathbb{E}\left[\hat{\omega}_t^2 \mid T = t_1\right]^{-1/2}, \dots, \mathbb{E}\left[\hat{\omega}_t^2 \mid T = t_n\right]^{-1/2} \right).
\]
The vector \( z_n \in \mathbb{R}^n \) collects the residuals weighted by the estimated weighting functions
\[
z_n = \begin{pmatrix} z_1 \\ \vdots \\ z_n \end{pmatrix}, \quad \text{where} \quad z_i = \left( y_i - \hat{\mu}(x_i, t_i) \right) \hat{\omega}_i.
\]
The expectation-adjusted residual vector \( \overset{\circ}{z}_n \in \mathbb{R}^n \) is
\[
\overset{\circ}{z}_n = \begin{pmatrix} \overset{\circ}{z}_1 \\ \vdots \\ \overset{\circ}{z}_n \end{pmatrix}, \quad \text{where} \quad \overset{\circ}{z}_i = \mathbb{E}\left[ (Y - \hat{\mu}_i(T, x_i)) \hat{\omega}_i \mid T = t_i \right].
\]
Considering the weighted loss function that incorporates targeted regularization, we have
\[
L = \sum_{i=1}^{n} \hat{\omega}_i \left( y_i - \hat{\mu}_t(x_i, t_i) - \hat{\omega}_i \epsilon(t_i) \right)^2.
\]
Minimizing this loss function yields the estimator for the coefficients \( \hat{\alpha} \)
\[
\hat{\alpha} = \left( \Phi_n \Lambda_n^{-3} \Phi_n^T \right)^{-1} \Phi_n \Lambda_n^{-1} z_n.
\]
The expectation form of \( \alpha \), denoted as \( \overset{\circ}{\alpha} \), is given by
\[
\mathring{\alpha} = \left( \Phi_n \Lambda_n^{-3} \Phi_n^T \right)^{-1} \Phi_n \Lambda_n^{-3} \mathring{\Lambda^2_n} \mathring{z}_n.
\]
For the purpose of analyzing function spaces and their complexities, let \( F_1 \) and \( F_2 \) denote function spaces. Suppose that for any function \( f \) in these spaces, the supremum norm satisfies \( \|f\|_\infty < \alpha \) and \( \|f\|_\infty < \infty \).

The Rademacher complexity of the union of these function spaces is bounded 
\[
\text{Rad}_n(F_1 \cup F_2) \leq \frac{1}{2} \left( \text{Rad}_n(F_1) + \text{Rad}_n(F_2) \right) \left( \|f_1\|_\infty + \|f_2\|_\infty \right).
\]
Here, \( \text{Rad}_n(F) \) represents the Rademacher complexity of the function space \( F \), defined by
\[
\text{Rad}_n(F) = \mathbb{E} \left( \sup_{f \in F} \frac{1}{n} \sum_{i=1}^{n} \sigma_i f(X_i) \right),
\]
where \( \sigma_i \) are independent Rademacher random variables.

\textbf{Lemma 2\ \ \ } The distance covariate optimal weight uniformly converges to the true balancing weights \( w = \frac{f(T)f(\mathbf{X})}{f(T,\mathbf{X})} \). 
 \[
 \lim_{n\to\infty} w_n = w
 \]

These results follow from a general framework for ADRF generalization across source and target populations. Setting the source and target to be the same recovers the result as a special case, corresponding to Theorem 3.8 of \cite{cheng2023causal}.

\textbf{Lemma 3\ \ \ }
$
\left\| \hat{\epsilon}_n(t) - \tilde{\epsilon}_n(t) \right\|_{L^2} = O_p\left(n^{-\frac{1}{3}} \sqrt{\log n}\right) \ \  where \ \ \tilde{\epsilon}_n(t) = \frac{\mathbb{E}\left[ (Y - \hat{\mu}_n(X, T)) \hat{\omega}_n \mid T = t \right]}{\mathbb{E}\left[\hat{\omega}_n^2 \mid T = t\right]}.
$

\begin{proof}
We proceed to establish the bounds required for Lemma 3 by decomposing the difference between the estimated residuals \( \hat{\epsilon}_n(t) \) and the expectation-adjusted residuals \( \tilde{\epsilon}_n(t) \) using the triangle inequality  \citep{nie2021vcnet}. We define \(\mathring{\epsilon}_n(t) = \sum_{k=1}^{K_n} \mathring{\alpha}_k \phi_k(t)\). We have
\[
\| \hat{\epsilon}_n(t) - \tilde{\epsilon}_n(t) \|_2 \leq \| \hat{\epsilon}_n(t) - \mathring{\epsilon}_n(t) \|_2 + \| \mathring{\epsilon}_n(t) - \tilde{\epsilon}_n(t) \|_2.
\]

\textbf{Bounding \( \| \hat{\epsilon}_n(t) - \mathring{\epsilon}_n(t) \|_2 \)}

Utilizing the boundedness of the B-spline basis functions, we derive
\[
\| \hat{\epsilon}_n(t) - \mathring{\epsilon}_n(t) \|_2 \leq C \frac{ \| \hat{\alpha} - \mathring{\alpha} \|_2 }{ \sqrt{K_n} },
\]
where \( C \) is a constant arising from the boundedness of the B-spline basis.

To bound \( \| \hat{\alpha} - \mathring{\alpha} \|_2 \). We invoke Rademacher complexity properties for product function classes. Let \(\mathbb{Q}\) and \(\mathbb{U}\) represent the functional spaces for weight functions \(w\) and outcome models \(\mu\), respectively. We have the bound
\[
\text{Rad}_n((\mathbb{U} + \mu)\mathbb{Q}) \leq \frac{1}{2}\left(\|\mathbb{U}\|_\infty + \|\mathbb{Q}\|_\infty\right) \left(\text{Rad}_n(\mathbb{U}) + \text{Rad}_n(\mathbb{Q})\right).
\]
Expanding the complexity of \(\mathbb{Q}\) via the Lipschitz composition property, step (a) follows from plugging \(h : x \mapsto \frac{1}{x - 1/2c} + 2c\) into Theorem 12(4) of Bartlett \& Mendelson (2002)
\begin{align*}
\text{Rad}_n((\mathbb{U} + \mu)\mathbb{Q}) &\stackrel{(a)}{\leq} \frac{1}{2}\left(\|\mathbb{U}\|_\infty + \|\mathbb{Q}\|_\infty\right) \left( \text{Rad}_n(\mathbb{U}) + \max\left( \frac{c^2}{2}, \frac{2}{(c - 1/c)^2} \right) \text{Rad}_n\left(\mathbb{Q} - \frac{1}{2c}\right) + \frac{2c}{n} \right) \\
&= O(n^{-1/2}).
\end{align*}

Defining the composite class \(\mathcal{A} = (\mathbb{U} + \mu)\mathbb{Q}\), we obtain
\[
\text{Rad}_n(\phi_k \mathcal{A}) \leq \frac{1}{2}\left( \|\phi_k\|_\infty + \|\mathcal{A}\|_\infty \right) \left( \text{Rad}_n(\phi_k) + \text{Rad}_n(\mathcal{A}) \right) = O(n^{-1/2}).
\]
We bound the first term of the probability decomposition using these results
\begin{align*}
\text{P} \left( \sup_{\hat{w}, \hat{\mu}} \left| \frac{1}{n} \sum_{i=1}^{n} \phi_k(t_i) u_i \right| > \frac{1}{2} \sqrt{\frac{a}{K_n}} \right) 
&\stackrel{(a)}{\leq} \frac{\mathbb{E} \left( \sup_{\hat{w}, \hat{\mu}} \left| \frac{1}{n} \sum_{i=1}^{n} \phi_k(t_i) u_i \right| \right)}{\frac{1}{2} \sqrt{\frac{a}{K_n}}} \\
&\stackrel{(b)}{\asymp}  \sqrt{\frac{K_n}{an}},
\end{align*}
where (a) follows from Markov's Inequality, and (b) utilizes the deﬁnition of Rademacher complexity.

We now control the second summation involving the noise term $\mathring{v}_i$. Employing a truncation strategy with a threshold $M_n > 0$, we partition the probability based on whether the noise magnitude $|v_i|$ exceeds this level
\begin{align*}
\text{P} \left( \sup_{\hat{w}, \hat{\mu}} \left| \frac{1}{n} \sum_{i=1}^{n} \phi_k(t_i) \mathring{v}_i \right| > \frac{1}{2} \sqrt{\frac{a}{K_n}} \right) 
\leq & \ \text{P} \left( \sup_{\hat{w}, \hat{\mu}} \left| \frac{1}{n} \sum_{i=1}^{n} \phi_k(t_i) \mathring{v}_i \mathbb{I}(|v_i| \leq M_n) \right| > \frac{1}{4} \sqrt{\frac{a}{K_n}} \right) \\
& + \text{P} \left( \sup_{\hat{w}, \hat{\mu}} \left| \frac{1}{n} \sum_{i=1}^{n} \phi_k(t_i) \mathring{v}_i \mathbb{I}(|v_i| > M_n) \right| > \frac{1}{4} \sqrt{\frac{a}{K_n}} \right).
\end{align*}
For the bounded component ($|v_i| \leq M_n$), applying Markov's inequality followed by the Rademacher complexity bound for bounded classes yields
\[
\text{P} \left( \sup_{\hat{w}, \hat{\mu}} \left| \frac{1}{n} \sum_{i=1}^{n} \phi_k(t_i) \mathring{v}_i \mathbb{I}(|v_i| \leq M_n) \right| > \frac{1}{4} \sqrt{\frac{a}{K_n}} \right)
\lesssim \frac{\mathbb{E} \sup_{\hat{w}, \hat{\mu}} \left| \frac{1}{n} \sum_{i=1}^{n} \phi_k(t_i) \hat{w}_i v_i \mathbb{I}(|v_i| \leq M_n) \right|}{\sqrt{a/K_n}} 
\lesssim \sqrt{\frac{K_n}{an}} M_n.
\]
For the tail component ($|v_i| > M_n$), we define the random variable representing the tail noise magnitude as \( W = |v|\mathbb{I}(|v| > M_n) \). Since the basis functions $\phi_k$ and weights $\hat{w}$ are uniformly bounded, the expectation of the supremum term is dominated by the expectation of \(W\). Applying Markov's inequality, we derive the bound 
\begin{align*}
\text{P} \left( \sup_{\hat{w}, \hat{\mu}} \left| \frac{1}{n} \sum_{i=1}^{n} \phi_k(t_i) \mathring{v}_i \mathbb{I}(|v_i| > M_n) \right| > \frac{1}{4} \sqrt{\frac{a}{K_n}} \right)
&\lesssim \frac{\mathbb{E} \sup_{\hat{w}, \hat{\mu}} \left| \frac{1}{n} \sum_{i=1}^{n} \phi_k(t_i) \hat{w}_i v_i \mathbb{I}(|v_i| > M_n) \right|}{\sqrt{a/K_n}} \\
&\stackrel{(a)}{\lesssim} \frac{\int_0^\infty (1 - F_W(w)) dw - \int_{-\infty}^0 F_W(w) dw}{\sqrt{a/K_n}}  \\
&= \frac{\int_0^\infty \text{P}(|v| \geq \max(M_n, w)) dw}{\sqrt{a/K_n}} \\
&\stackrel{(b)}{\lesssim} \frac{\int_0^\infty e^{-\sigma [\max(M_n, w)]^2} dw}{\sqrt{a/K_n}} \\
&\leq \frac{\int_0^\infty e^{-\sigma [M_n + w]^2} dw}{\sqrt{a/K_n}} \\
&\stackrel{(c)}{\lesssim} \frac{e^{-\sigma M_n^2}}{M_n} \frac{\sqrt{K_n}}{\sqrt{a}},
\end{align*}
where (a) uses the integral formula for expectation \(\mathbb{E}W = \int_0^\infty (1-F(w))dw - \int_{-\infty}^0 F(w)dw\), (b) utilizes the fact that \(v\) follows a sub-Gaussian distribution, and (c) applies Mills' ratio approximation.

To optimize the bound, we set \( M_n \asymp \sqrt{\log n} \) and \( a \asymp \frac{ K_n \log n }{ n } \), which yields (assuming the first term of $\| \hat{\alpha} - \mathring{\alpha} \|_2$ dominates or the second term is of similar or smaller order)
\[
\left\| \left( \Phi_n \Pi_n^{-2} \Phi_n^T \right)^{-1} \Phi_n^T (Z_n - \tilde{Z}_n) \right\|_2 = O_p\left( \sqrt{ \frac{ K_n^3 \log n }{ n } } \right).
\]
Similarly, we have
\[
\| \hat{\alpha} - \mathring{\alpha} \|_2 = O_p\left( \sqrt{ \frac{ K_n^3 \log n }{ n } } \right).
\]
Substituting back into the bound for the first part, we obtain
\[
\| \hat{\epsilon}_n(t) - \mathring{\epsilon}_n(t) \|_2 = O_p\left( \sqrt{ \frac{ K_n^2 \log n }{ n } } \right).
\]

\textbf{Second Part: Bounding \( \|\mathring{\epsilon}_n(t) - \tilde{\epsilon}_n(t)\|_2 \)}

Let \(\check{\alpha} \in \mathbb{R}^{K_n}\) be such that \(\| (\check{\alpha})^T \phi^{K_n} - \tilde{\epsilon}_n \|_\infty = \inf_{f \in \mathcal{B}_{K_n}} \| f - \tilde{\epsilon}_n \|_\infty\). Applying the triangle inequality, we obtain
\begin{align*}
\| \mathring{\epsilon}_n - \tilde{\epsilon}_n \|_{L^2} 
&\leq \| \mathring{\epsilon}_n - (\check{\alpha})^T \phi^{K_n} \|_{L^2} + \| (\check{\alpha})^T \phi^{K_n} - \tilde{\epsilon}_n \|_{L^2}.
\end{align*}

By the definition of \(\check{\alpha}\) and the properties of the B-spline space and the assumption 2, we have a bound on the second term
\[
\| \tilde{\epsilon}_n(t) - \tilde{\alpha}^\top \phi^{K_n} \|_2 = O_p \left( \inf_{f \in \mathrm{span}\{\phi^{K_n}\}} \sup_t \| \tilde{\epsilon}_n(t) - f(t) \| \right) = O_p(K_n^{-2}).
\]
Notice that the first term can also be bounded. Recalling that \(\mathring{\epsilon}_n(t) = \mathring{\alpha}^T \phi^{K_n}(t)\), we have
\begin{align*}
\| (\check{\alpha})^T \phi^{K_n} - \mathring{\epsilon}_n \|_{L^2} 
&\stackrel{(a)}{\leq} \| \mathring{\alpha} - \check{\alpha} \|_2 / \sqrt{K_n} \\
&= \left\| \left( \Phi_n \Lambda_n^{-3} \Phi_n^T \right)^{-1} \Phi_n \Lambda_n^{-1} \left( \Phi_n \check{\alpha} - \Lambda_n^{-2} \mathring{\Lambda}_n^2 \mathring{z}_n \right) \right\|_2 / \sqrt{K_n} \\
&\asymp \frac{K_n}{n} \left\| \Phi_n \Lambda_n^{-1} \left( \Phi_n \check{\alpha} - \Lambda_n^{-2} \mathring{\Lambda}_n^2 \mathring{z}_n \right) \right\|_2 / \sqrt{K_n} \\
&\stackrel{(b)}{\asymp} \frac{K_n^{-1.5}}{n} \sqrt{ \mathbf{1}^T \Lambda_n^{-1} \Phi_n^T \Phi_n \Lambda_n^{-1} \mathbf{1} } \\
&\asymp \frac{K_n^{-1.5}}{n} \sqrt{ \sum_{k=1}^{K_n} \left( \sum_{i=1}^n \phi_k(t_i)\hat{w}_i \right)^2 } \\
&\asymp K_n^{-1.5} \sqrt{ \sum_{k=1}^{K_n} \left( \frac{1}{n} \sum_{i=1}^n \phi_k(t_i) \right)^2 },
\end{align*}
where (a) follows from the properties of B-spline basis functions, and (b) follows from the properties of the B-spline space such that \(\| \Phi_n \check{\alpha} - \Lambda_n^{-2} \mathring{\Lambda}_n^2 \mathring{z}_n \|_\infty = O_p(K_n^{-2})\) because \((\tilde{\epsilon}_n(t_1), \dots, \tilde{\epsilon}_n(t_n))^T = \Lambda_n^{-2} \mathring{\Lambda}_n^2 \mathring{z}_n\). 

Following the proof of Lemma A.6 of \citep{huang2004polynomial}, for any \(a > [\mathbb{E}\phi_k(T)]^2 K_n\), we have
\begin{align*}
\text{Prob} \left( \sum_{k=1}^{K_n} \left( \frac{1}{n} \sum_{i=1}^n \phi_k(t_i) \right)^2 > a \right) 
&\stackrel{(a)}{\leq} \sum_{k=1}^{K_n} \text{Prob} \left( \left| \frac{1}{n} \sum_{i=1}^n \phi_k(t_i) \right| > \sqrt{\frac{a}{K_n}} \right) \\
&\leq \sum_{k=1}^{K_n} \text{Prob} \left( \left| \frac{1}{n} \sum_{i=1}^n \phi_k(t_i) - \mathbb{E}\phi_k(T) \right| > \sqrt{\frac{a}{K_n}} - |\mathbb{E}\phi_k(T)| \right) \\
&\stackrel{(b)}{\leq} 2K_n \exp \left\{ -2n \left( \sqrt{a/K_n} - |\mathbb{E}\phi_k(T)| \right)^2 \right\},
\end{align*}
where (a) uses the union bound, and (b) follows from Hoeffding's Inequality for bounded random variables. Since \(\mathbb{E}\phi_k(T) \asymp 1/K_n\), we can pick \(a = 2[\mathbb{E}\phi_k(T)]^2 K_n \asymp 1/K_n\), and thus \(\sum_{k=1}^{K_n} \left( \frac{1}{n} \sum_{i=1}^n \phi_k(t_i) \right)^2 = O_p(1/K_n)\). Plugging this into the previous equation, we get
\[
\| (\check{\alpha})^T \phi^{K_n} - \mathring{\epsilon}_n \|_{L^2} = O_p(K_n^{-2}).
\]
Thus, we can bound the bias term
\[
\| \mathring{\epsilon}_n - \tilde{\epsilon}_n \|_{L^2} = O_p(K_n^{-2}).
\]

Combining the bounds for both terms, we finally obtain
\[
\| \tilde{\epsilon}_n(t) - \mathring{\epsilon}_n(t) \|_2 = O_p(K_n^{-2}).
\]
Combining both parts, the overall bound on the difference between the estimated residuals and the expectation-adjusted residuals is
\[
\| \hat{\epsilon}_n(t) - \tilde{\epsilon}_n(t) \|_{L^2} = O_p\left( K_n^{-2} + \frac{ K_n \sqrt{ \log n } }{ \sqrt{ n } } \right).
\]
Selecting \( K_n \asymp n^{1/6} \) balances the two terms, yielding the final bound
\[
\| \hat{\epsilon}_n(t) - \tilde{\epsilon}_n(t) \|_{L^2} = O_p\left( n^{-1/3} \sqrt{ \log n } \right).
\]
This concludes the proof of Lemma 3, establishing the necessary bounds for both components of the triangle inequality.
\end{proof}

\textbf{Proof of Theorem 1} 

\begin{proof}
We begin by expressing the difference between the estimated function \( \hat{\varphi}(t) \) and the true function \( \varphi(t) \)
\[
\hat{\varphi}(t) = \frac{1}{n} \sum_{i=1}^{n} \left( \hat{\mu}(t, \boldsymbol{x}_i) + \hat{\epsilon}(t) \hat{w}_i \right) = \frac{1}{n} \sum_{i=1}^{n} \hat{\mu}(t, \boldsymbol{x}_i) + \hat{\epsilon}(t),
\]

The \( L^2 \) norm of their difference is given by
\[
\|\hat{\varphi}(t) - \varphi(t)\|_{L^2}^2 = \int \left| \hat{\varphi}(t) - \varphi(t) \right|^2 dt.
\]

\[
\begin{aligned}
\|\hat{\varphi}(t) - \varphi(t)\|_{L^2} &\leq \left\| \hat{\epsilon}(t) - \mathbb{E}\left(Y - \hat{\mu}_n(T, \boldsymbol{X}) \hat{w}_n \mid T = t\right) \right\|_{L^2}  + \left\| \frac{1}{n} \sum_{i=1}^{n} \hat{\mu}_n(t, \boldsymbol{x}_i) - \mathbb{E}\left(\hat{\mu}_n(t, \boldsymbol{X})\right) \right\|_{L^2} \\
& \quad + \left\| \mathbb{E}\left(Y - \hat{\mu}_n(T, \boldsymbol{X}) \hat{w}_n \mid T = t\right) + \mathbb{E}\left(\hat{\mu}_n(t, \boldsymbol{X}) \right) - \varphi(t) \right\|_{L^2}.
\end{aligned}
\]

Next, we define the estimator \( \tilde{\epsilon}(t) \) as
\[
\tilde{\epsilon}(t) = \frac{\mathbb{E}\left[ (Y - \hat{\mu}_n(\boldsymbol{X}, T)) \hat{w}_n \mid T = t \right]}{\mathbb{E}\left[\hat{w}_n^2 \mid T = t\right]}.
\]

We then bound the first term
\[
\begin{aligned}
&\left\| \hat{\epsilon}(t) - \mathbb{E}\left(Y - \hat{\mu}_n(T, \boldsymbol{X}) \hat{w}_n \mid T = t\right) \right\|_{L^2} \\
&\leq \left\| \hat{\epsilon}(t) - \tilde{\epsilon}(t) \right\|_{L^2} \\
&\quad + \left\| \tilde{\epsilon}(t) \int \hat{w}_n \, d(\hat{F}_n(x) - F(x)) \right\|_{L^2} \\
&\quad + \left\| \tilde{\epsilon}(t) \int \hat{w}_n \, dF(x) - \mathbb{E}\left(Y - \hat{\mu}_n(T, \boldsymbol{X}) \hat{w}_n \mid T = t\right) \hat{w}_n \right\|_{L^2}.
\end{aligned}
\]

Simplifying the last term, we obtain
\[
\begin{aligned}
 \left\| \hat{\epsilon}(t) - \tilde{\epsilon}(t) \right\|_{L^2}  + \left\| \tilde{\epsilon}(t) \int \hat{w}_n \, d(\hat{F}_n(x) - F(x)) \right\|_{L^2} + \left\| \tilde{\epsilon}(t) (\int \hat{w}_n \, dF(x) - \int \frac{\hat{w}_n^2}{w} \, dF(x)) \right\|_{L^2}.
\end{aligned}
\]

Further simplifying using the properties of expectations and integrals
\[
\begin{aligned}
&= \left\| \mathbb{E}\left( \hat{\mu}(\boldsymbol{X},T) - \mu(T, \boldsymbol{x}) \hat{w}_n \mid T = t \right) \int \hat{w}_n\left(1 - \frac{\hat{w}_n}{w}\right) dF(x) \bigg/ \mathbb{E}\left(\hat{w}_n^2\right) \right\|_{L^2}.
\end{aligned}
\]

From the relevant lemma, we have
\[
\left\| \hat{\epsilon}(t) - \tilde{\epsilon}(t) \right\|_{L^2} = O_p\left(n^{-\frac{1}{3}} \sqrt{\log n}\right).
\]

Next, consider the second term involving \( \hat{\mu}_n(t, \boldsymbol{x}_i) \):
From the generalization bound and assumption, we know that
\[
\sup_{t_0 \in [0,1]} \left| \frac{1}{n} \sum_{i=1}^{n} \hat{\mu}_n(\boldsymbol{x}_i, t_0) - \mathbb{E}\left(\hat{\mu}_n(t_0, \boldsymbol{X})\right) \right| = O_p\left(n^{-\frac{1}{2}}\right).
\]
Thus, in the \( L^2 \) norm
\[
\left\| \frac{1}{n} \sum_{i=1}^{n} \hat{\mu}_n(\boldsymbol{x}_i, \cdot) - \mathbb{E}\left(\hat{\mu}_n(t, \boldsymbol{X})\right) \right\|_{L^2} = O_p\left(n^{-\frac{1}{2}}\right).
\]

Recall Lemma 1, which states that if
\[
\sup_{t \in [0,1]} \sup_{\boldsymbol{X} \in \mathcal{X}} \left| \hat{w}_n(t,X) - w(t,X) \right| = O_p\left(r_2(n)\right),
\]
\[
\sup_{t \in [0,1]} \sup_{\boldsymbol{X} \in \mathcal{X}} \left| \hat{\mu}_n(t, \boldsymbol{X}) - \mu(t, \boldsymbol{X}) \right| = O_p\left(r_1(n)\right),
\]
\textit{then}
\[
\sup_{t_0 \in [0,1]} \left| \mathbb{E} \left[ \delta(T - t_0) (Y - \hat{\mu}_n(T, \boldsymbol{X}))\hat{w}_n + \hat{\mu}_n(t_0, \boldsymbol{X}) \right] - \psi(t_0) \right| = O_p\left(r_1(n) r_2(n)\right).
\]

Combining all bounded terms, we obtain
\[
\begin{aligned}
\|\hat{\varphi}(t) - \varphi(t)\|_{L^2} &\leq \left\| \hat{\epsilon}(t) - \mathbb{E}\left(Y - \hat{\mu}_n(T, \boldsymbol{X}) \hat{w}_n \mid T = t\right) \right\|_{L^2} \\
&\quad + \left\| \frac{1}{n} \sum_{i=1}^{n} \hat{\mu}_n(t, \boldsymbol{x}_i) - \mathbb{E}\left(\hat{\mu}_n(t, \boldsymbol{X})\right) \right\|_{L^2} \\
&\quad + \left\| \mathbb{E}\left(Y - \hat{\mu}_n(T, \boldsymbol{X}) \hat{w}_n \mid T = t\right) + \frac{1}{n} \sum_{i=1}^{n} \hat{\mu}_n(t, \boldsymbol{x}_i) - \varphi(t) \right\|_{L^2} \\
&= O_p\left(n^{-\frac{1}{3}} \sqrt{\log n} + (r_1(n) r_2(n)\right) + O_p\left(n^{-\frac{1}{2}}\right) + O_p\left(r_1(n) r_2(n)\right).
\end{aligned}
\]

We conclude
\[
\|\hat{\varphi}(t) - \varphi(t)\|_{L^2} = O_p\left(n^{-\frac{1}{3}} \sqrt{\log n} + r_1(n) r_2(n)\right).
\]

This completes the proof of Theorem 1.
\end{proof}

\section{Dataset and Experiment Setting}
\subsection{Experiment Setting Details}
We set the training parameters based on the size of each dataset. For IHDP, we use 20 replicates with 800 training epochs and a learning rate of 0.0005. For the News dataset, we use 10 replicates with 600 epochs and the same learning rate. For the TCGA dataset, we apply 5 replicates with 1000 epochs and a smaller learning rate of 0.00005. Across all methods, we choose the number of grid points which are all equally spaced at $[0,1]$ from \{10, 14, 18, 22\} and the activation function from \{ReLU, tanh, sigmoid\}. A B-spline basis with degree 2 is used throughout. To ensure fair comparison, we keep the set of hyperparameters consistent across different deep learning methods within the same dataset. Experiments were performed on a Macbook Air with M1 chip and 16 GB of RAM and high-performance computing (HPC) cluster.

\subsubsection{IHDP}
The original semi-synthetic IHDP dataset contains binary treatments with 747 observations on 25 covariates. To allow comparison on continuous treatments, we randomly generated treatments and responses using the following equations.

\begin{align*}
\widetilde{t} \mid \boldsymbol{x} &= \frac{2x_1}{(1+x_2)} + \frac{2\max(x_3,x_5,x_6)}{0.2+\min(x_3,x_5,x_6)} + 2\tanh\left(5 \frac{\sum_{i \in S_{\mathrm{dis},2}}(x_i - c_2)}{|S_{\mathrm{dis},2}|}\right) - 4 + \mathcal{N}(0, 0.25), \\
y \mid \boldsymbol{x}, t &= \frac{\sin(3\pi t)}{1.2 - t} \left( \tanh\left(5 \frac{\sum_{i \in S_{\mathrm{dis},1}}(x_i - c_1)}{|S_{\mathrm{dis},1}|}\right) + \frac{\exp(0.2(x_1 - x_6))}{0.5 + 5 \min(x_2, x_3, x_5)} \right) + \mathcal{N}(0, 0.25),
\end{align*}

where $t = (1 + \exp(-\tilde{t}))^{-1}$, $S_{\mathrm{con}} = \{1, 2, 3, 5, 6\}$ is the index set of continuous features, and the discrete feature sets are $S_{\mathrm{dis},1} = \{4, 7, 8, 9, 10, 11, 12, 13, 14, 15\}$ and $S_{\mathrm{dis},2} = \{16, 17, 18, 19, 20, 21, 22, 23, 24, 25\}$.

The constants $c_1$ and $c_2$ are defined as
\begin{align*}
c_1 &= \mathbb{E} \left( \frac{\sum_{i \in S_{\mathrm{dis},1}} x_i}{|S_{\mathrm{dis},1}|} \right), \quad
c_2 = \mathbb{E} \left( \frac{\sum_{i \in S_{\mathrm{dis},2}} x_i}{|S_{\mathrm{dis},2}|} \right).
\end{align*}

\subsubsection{News}
The \textbf{News} dataset consists of 3,000 randomly sampled news items with 500 covariates from the NY Times corpus. We first generated the vectors $\boldsymbol{v}_1^{\prime}, \boldsymbol{v}_2^{\prime}$, and $\boldsymbol{v}_3^{\prime}$ from $\mathcal{N}(\mathbf{0}, \mathbf{1})$. Then, we normalized them by setting:

\[
\boldsymbol{v}_i = \frac{\boldsymbol{v}_i^{\prime}}{\|\boldsymbol{v}_i^{\prime}\|_2}, \quad \text{for } i = 1, 2, 3.
\]

Given the covariates $\boldsymbol{x}$, we generated the treatment variable $t$ from a Beta distribution
\[
t \sim \text{Beta}\left(2, \left|\frac{\boldsymbol{v}_3^\top \boldsymbol{x}}{2 \boldsymbol{v}_2^\top \boldsymbol{x}}\right|\right).
\]

The outcome variable $y$ was generated in two steps
\[
y^{\prime} \mid \boldsymbol{x}, t = \exp\left(\frac{\boldsymbol{v}_2^\top \boldsymbol{x}}{\boldsymbol{v}_3^\top \boldsymbol{x}} - 0.3\right),
\]
\[
y \mid \boldsymbol{x}, t = 2\left(\max(-2, \min(2, y^{\prime})) + 20 \boldsymbol{v}_1^\top \boldsymbol{x}\right) \cdot \left(4\left(t - 0.5\right)^2 \cdot \sin\left(\frac{\pi}{2} t\right)\right) + \mathcal{N}(0, 0.5).
\]

\subsubsection{TCGA}
The \textbf{TCGA} dataset comprises 9659 observations, each with 4000 covariates. The dataset also has continuous treatment t. The outcome $y$ is generated by first sampling a set of parameters $\mathbf{v}_1$, $\mathbf{v}_2$, and $\mathbf{v}_3$. These parameters are obtained by drawing a vector $\mathbf{u}$ from $\mathcal{N}(\mathbf{0}, \mathbf{1})$ and setting $\mathbf{v} = \mathbf{u}/\|\mathbf{u}\|$, where $\|\cdot\|$ denotes the Euclidean norm. The outcome is then generated as $$y = f(\mathbf{x}, t) = (\mathbf{v}_1^2)^T\mathbf{x} + \sin\left(\pi\left(\frac{(\mathbf{v}_2^2)^T\mathbf{x}}{\mathbf{v}_3^T\mathbf{x}}\right) t\right)$$.

\end{document}

%% file: math_commands.tex
\usepackage{amsmath,amsfonts,bm}

\def\eqref#1{equation~\ref{#1}}

\def\1{\bm{1}}

\DeclareMathAlphabet{\mathsfit}{\encodingdefault}{\sfdefault}{m}{sl}
\SetMathAlphabet{\mathsfit}{bold}{\encodingdefault}{\sfdefault}{bx}{n}

